\documentclass{article}
\usepackage{arxiv}
\usepackage{times}
\usepackage[round,authoryear]{natbib}

\usepackage{amsmath,amsfonts,bm}

\def\eqref#1{equation~\ref{#1}}
\def\1{\bm{1}}

\DeclareMathAlphabet{\mathsfit}{\encodingdefault}{\sfdefault}{m}{sl}
\SetMathAlphabet{\mathsfit}{bold}{\encodingdefault}{\sfdefault}{bx}{n}

\usepackage[utf8]{inputenc} % allow utf-8 input
\usepackage[T1]{fontenc}    % use 8-bit T1 fonts
\usepackage{hyperref}       % hyperlinks
\usepackage{url}            % simple URL typesetting
\usepackage{booktabs}       % professional-quality tables
\usepackage{amsfonts}       % blackboard math symbols
\usepackage{nicefrac}       % compact symbols for 1/2, etc.
\usepackage{microtype}      % microtypography
\usepackage{xcolor}         % colors
\usepackage{graphicx}
\usepackage{float}
\usepackage{subcaption}
\usepackage{algorithm}
\usepackage{algpseudocode}
\usepackage{amsmath}
\usepackage{tikz}
\usetikzlibrary{arrows.meta,calc,positioning,fit,backgrounds}
\usepackage{array}
\newcolumntype{H}{>{\setbox0=\hbox\bgroup}c<{\egroup}@{}}

\usepackage[T1]{fontenc}
\usepackage{pgfplots}
\pgfplotsset{compat=1.18}
\usepackage{subcaption}

\usepackage{theorem}
\usepackage{fancybox,amssymb,amstext,amsmath,stmaryrd,wasysym,proof,mathtools,multirow,stackrel}
\SetSymbolFont{stmry}{bold}{U}{stmry}{m}{n}
\SetSymbolFont{wasy}{bold}{U}{wasy}{m}{n}
\usepackage{algorithm}
\usepackage{algpseudocode}
\makeatletter
\algrenewcommand\ALG@beginalgorithmic{\footnotesize}
\makeatother

\usepackage{booktabs}
\usepackage{longtable}
\usepackage{array}

\theorembodyfont{\itshape}
\newtheorem{mytheorem}{Theorem}[section]
\newtheorem{mylemma}[mytheorem]{Lemma}
\newtheorem{myproposition}[mytheorem]{Proposition}

\newtheorem{mycorollary}[mytheorem]{Corollary}
\theorembodyfont{\rmfamily}

\newtheorem{myremark}[mytheorem]{Remark}
\newtheorem{myexample}[mytheorem]{Example}
\newtheorem{mydefinition}[mytheorem]{Definition}

\newenvironment{proof}{\par\noindent\textit{Proof. }\ignorespaces}{\hfill$\square$\par}

\newcommand{\ttrue}{\mathrm{t{\kern-1.5pt}t}}
\newcommand{\ffalse}{\mathrm{f{\kern-1.5pt}f}}

\newcommand{\UntilOp}[1]{\mathbin{\mathcal{U}}}
\newcommand{\Release}[1]{\mathbin{\mathcal{R}}}

\newcommand{\DiaOp}[1]{\Diamond}
\newcommand{\BoxOp}[1]{\square}

\renewcommand{\paragraph}[1]{\noindent \textbf{#1}}

\title{Temporal Specification Guidance for Text-to-Video Generation}

\author{
Avishree Khare\textsuperscript{1},
Hideki Okamoto\textsuperscript{2},
Bardh Hoxha\textsuperscript{2},
Georgios Fainekos\textsuperscript{2},
Rajeev Alur\textsuperscript{1} \\
\textsuperscript{1}University of Pennsylvania \\
\textsuperscript{2}Toyota Motor North America (R\&D)
}

\begin{document}

\maketitle

\begin{abstract}
Text-to-video models often generate the visual concepts requested in natural language prompts but fail to satisfy their intended ordering, timing, and persistence requirements. We propose representing these requirements as specifications in a formal Temporal Logic, and introduce \textbf{Temporal Specification Guidance (TSG)}, a training-free inference-time framework that turns temporal specifications into executable objectives for guiding video generation during sampling. TSG scores traces of visual concept evidence using differentiable quantitative semantics and uses the score's gradients to construct a specification-guided denoising direction. We instantiate this framework with video--text attention traces, efficient differentiable scores for Metric Temporal Logic, and gradient-derived attention biases. This instantiation requires neither an external concept classifier nor backpropagation through the video transformer. Its attention-steered forward pass yields a denoising direction that augments classifier-free guidance (CFG). On a benchmark of 220 prompts spanning 11 categories of temporal constraints, including ordered object entry and state transitions, this instantiation improves specification satisfaction accuracy over CFG by 9.5--20 percentage points across five models, with minimal memory overhead. These results demonstrate the potential of temporal specifications and differentiable satisfaction objectives as a flexible interface for controlling video generation.
\end{abstract}

\section{Introduction}

\begin{figure}[t]
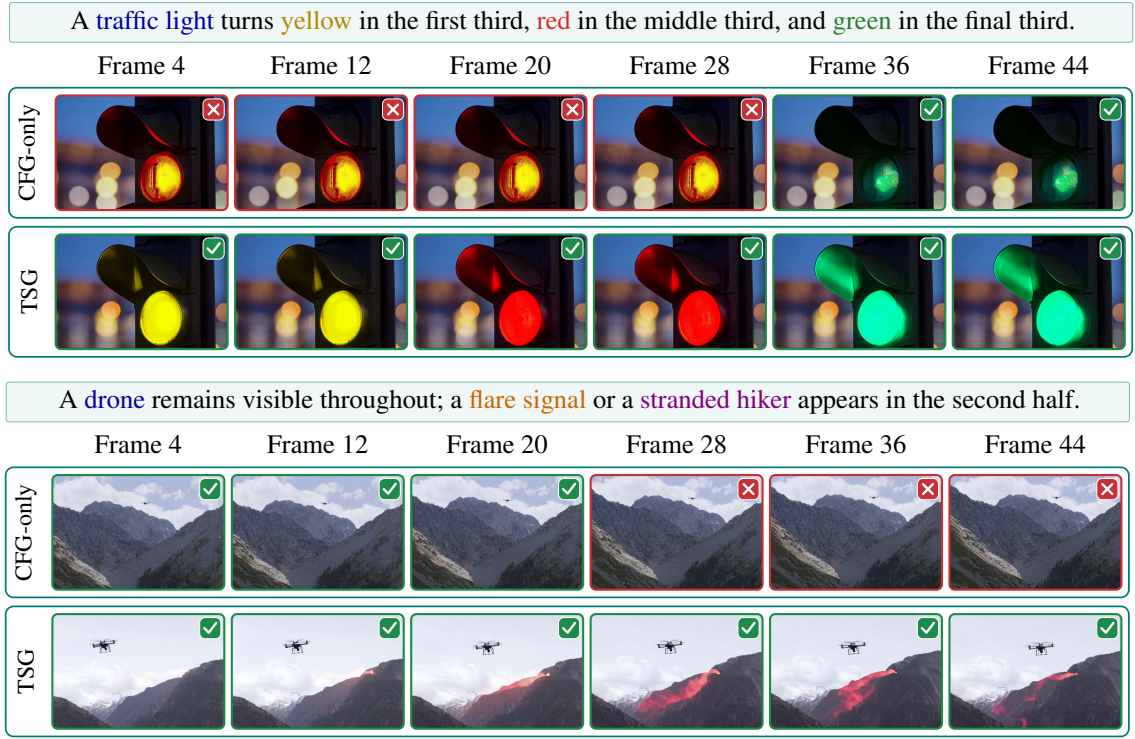

      \centering

      \begin{subfigure}{\textwidth}
          \centering
          \input{ICLR_2027/figures/examples/traffic_light/figure}

      \end{subfigure}

      \vspace{0.8em}

      \begin{subfigure}{\linewidth}
          \centering
         \input{ICLR_2027/figures/examples/drone/figure}
          \label{fig:motivation-storm-rainbow}
      \end{subfigure}

      \caption{TSG improves adherence to temporal requirements. \textbf{Top:} CFG produces an amber light for the first two-thirds while TSG follows the expected yellow--red--green progression. \textbf{Bottom:} CFG omits both requested rescue cues; TSG generates a flame signal while retaining the drone. Videos are generated with CogVideoX-5B.}
      \vspace{-0.2in}
      \label{fig:motivating-examples}
  \end{figure}

Text-to-Video generation has seen tremendous progress in recent years, with models such as Gemini Omni~\citep{deepmind2026geminiomni},  Runway Gen~\citep{runwaygen1,runway2025gen45}, 
% Pika~\citep{pika25},
Wan~\citep{wan2025wan}, and CogVideoX~\citep{yang2025cogvideox} producing videos with improved visual quality and semantic fidelity. Classifier-Free Guidance~\citep{ho2022cfg} was introduced to improve adherence to natural language prompts in image generation and is now widely adopted by video generation models~\citep{ho2022video,ho2022imagen}.
However, adherence to temporal constraints remains challenging~\citep{choi2025fixinpost,schiber2026tempocontrol}. Figure~\ref{fig:motivating-examples} shows two examples of such failures with CogVideoX-5B. While the model broadly incorporates the semantic concepts described in the text prompts (e.g., the presence of a traffic light or a drone), it fails to follow the required ordering, persistence, and timing requirements. In these examples, the model misses the intended timed color progression and the eventual appearance of a hiker or a flare signal, respectively.

In this work, we propose expressing temporal requirements as specifications in formal Temporal Logics. Formal temporal specifications can precisely express requirements that are implicitly described in natural language prompts. Frame-level visual concepts, such as the presence of a traffic light or the associated colors, can be represented as atomic propositions. Temporal Logics offer standard Boolean operators (negation, conjunction, and disjunction) and temporal operators such as Always ($\mathbf{G}$), Until ($\mathbf{U}$), and Eventually ($\mathbf{F}$) to express persistence and ordering requirements. In this work, we present a specification language adopted from Metric Temporal Logic~\citep{mtl} to express temporal requirements with different degrees of timing precision. Representations of temporal requirements from prior work, including timed binary masks~\citep{schiber2026tempocontrol} and untimed Temporal Logic formulae~\citep{choi2025fixinpost}, can also be written as specifications in this language.

We propose \textbf{Temporal Specification Guidance} (TSG), a training-free inference-time framework for improving adherence to temporal specifications with transformer-based video generation models. 
Our key insight is that temporal specifications provide an executable objective for guiding video generation during sampling. TSG makes this practical through differentiable specification scores and attention interventions that avoid backpropagation through the video transformer. Concretely, TSG computes traces of visual concept evidence using video-text attention from the video transformer. For every attention head, TSG computes biases as gradients of a differentiable specification score, steers the attention scores with these biases, and computes the specification-guided direction from the steered forward pass through the video transformer. 
Our use of attention as a proxy for visual concept presence builds on prior work on attention-based semantic guidance in image and video generation~\citep{attendandexcite,schiber2026tempocontrol}. 
The use of video-text attention mitigates the need for an external concept classifier during generation. Because differentiation is restricted to the attention-derived traces and their specification scores, TSG avoids backpropagation through the video transformer, reducing the associated computational and memory demands.

We compute a differentiable specification score over attention traces using a quantitative semantics for Metric Temporal Logic. Inspired by efficient log-domain quantitative semantics for Linear Temporal Logic~\citep{logstop}, our quantitative semantics aggregates support for the temporal witnesses and alternatives admitted by a specification. Its log-domain form provides differentiable guidance and can be efficiently evaluated in time linear in trace length for a fixed specification. We establish monotonicity in independent literal-support inputs and predictable concept-score gradient signs under consistent polarity.

We evaluate TSG on a benchmark of 220 prompts spanning temporal constraints from 11 categories, including timed state transitions, entity appearance and disappearance, persistence, and choice. Across five models from the CogVideoX~\citep{yang2025cogvideox} and Wan~\citep{wan2025wan} families, TSG improves specification satisfaction accuracy by 9.5--20 percentage points over CFG-only sampling. We also evaluate TSG with Wan 2.1 1.3B on strictly-timed prompts from the TempoControl object appearance benchmark~\citep{schiber2026tempocontrol}, where it outperforms CFG sampling by up to 22 percentage points in temporal accuracy and matches the state-of-the-art method on that benchmark with approximately 62\% lower peak GPU memory. Together, these findings support specification-guided attention steering as an effective and memory efficient approach to temporal control in video generation.

\section{Temporal constraints as formal specifications}
\label{sec:problem-setup}

Natural Language prompts to video generation models discuss the visual concepts that must be present in the video, along with the order and optionally the timing of their appearance. 
In this section, we discuss a formal specification language for these constraints.

Let \(\mathcal{M}\) represent a Text-to-Video (T2V) generation model that takes a text prompt \(p\) as input and outputs a video \(v = [v_1, \ldots, v_T]\) with \(T\) discrete frames. Let \(C\) denote the set of all frame-level visual concepts of interest. Examples of such visual concepts include entity presence (e.g., traffic light), entity states (e.g., the color red), and settings with properties  (e.g., sky at dusk). We use \(y(v_i, c) \in \{0, 1\}\) to denote the presence or absence of the visual concept \(c\) in frame \(v_i\). 

\paragraph{Specification language for temporal constraints.}
We propose a fragment of Metric Temporal Logic over finite traces (\(MTL_f\)) as a suitable language for specifying temporal constraints over visual concepts~\citep{mtl}. A specification \(\varphi\) in \(MTL_f\) takes the following form:

\[
\begin{aligned}
\varphi \coloneq{}& \top \mid \bot \mid c \mid \neg c
  \mid \varphi_1 \land \varphi_2 \mid \varphi_1 \lor \varphi_2 \mid \mathbf{X}_I \varphi \mid \varphi_1 \mathbf{U}_I \varphi_2 \mid \mathbf{G}_I \varphi
\end{aligned}
\]

where \(\top\) and \(\bot\) represent \textit{True} and \textit{False}, respectively, and \(c\) denotes an atomic proposition corresponding to a visual concept \(c\). The Boolean operators \(\neg\), \(\land\), and \(\lor\) denote negation, conjunction, and disjunction, respectively. The temporal operators \(\mathbf{X}_I\), \(\mathbf{U}_I\) and \(\mathbf{G}_I\) denote \textit{Next}, \textit{Until}, and \textit{Always}, in interval \(I \subseteq [0,1]\), resp. The temporal operator \textit{Eventually} \(\mathbf{F}_I\) can be derived as \(\mathbf{F}_I \varphi = \top \mathbf{U}_I \varphi\).  

This language can be used to specify several interesting constraints on persistence, order, and timing of visual concepts.
Time intervals \(I = [t_1, t_2]\) allow us to express specifications with varying timing precision:  untimed (\(t_1=0, t_2=1\)), interval-timed (\(t_1\ge0, t_2\le1\)), and strictly-timed (\(0 \le t_1=t_2\le1\)). Note that the intervals \(I=[t_1, t_2]\) are written over time normalized in \([0,1]\) and the corresponding offsets can be computed as \([\left\lceil t_1(T-1)\right\rceil, \left\lfloor t_2(T-1)\right\rfloor]\).
The constraint from Figure~\ref{fig:motivating-examples} that requires the drone to remain visible \textit{throughout the video} and at least one of the two rescue cues to appear \textit{at some point} in the second half can be written in this language as:
\[
\varphi_{\mathrm{rescue}}=
  \mathbf{G}_{[0, 1]}\mathsf{drone}
  \land \mathbf{F}_{[0.5,1]}
  (\mathsf{flare\_signal}\lor\mathsf{stranded\_hiker})
\]

\paragraph{Representing temporal constraints from prior work.} 
This specification language can be used to express temporal constraints from recent works, including untimed Temporal Logic (TL) specifications~\citep{choi2025fixinpost} and binary temporal masks that prescribe the presence or absence of a visual concept in each frame~\citep{schiber2026tempocontrol}. Untimed TL specifications can be written with the trivial \([0, 1]\) intervals. A temporal mask can be encoded as a conjunction of \emph{Always} constraints over intervals where the mask is constant, using \(c\) for presence and \(\neg c\) for absence. Additionally, we can write specifications that correspond to a set of temporal masks, including the choice between visual concepts  and the disappearance of an object within an interval. 
In other words, a temporal mask corresponds to a given valid trace while our specifications can admit several valid traces.

\paragraph{Deciding whether a video satisfies a specification.} These specifications can be supplied directly with the text prompt, or can be extracted from the prompt using a Large Language Model (LLM). Given the frame-level labels \(y(v_i, c)\) for visual concepts, the label \(y(v, \varphi, t)\) for a video \(v\) satisfying specification \(\varphi\) starting from frame \(1 \leq t \leq T\) can be computed deterministically using the standard Boolean semantics for \(MTL_f\). We present this semantics, along with examples of other specifications, in Appendix~\ref{appendix:mtl_semantics}. The examples in Figure~\ref{fig:motivating-examples} thus score \(y(v, \varphi, 1) = 0\) with CFG-only sampling, which we discuss next, and \(y(v, \varphi, 1) = 1\) with TSG, discussed in Section~\ref{sec:tsg}.

\section{Video Generation and Classifier Free Guidance}
\label{sec:background}

We focus on latent text-to-video models that generate videos from text prompts through iterative sampling in a compressed latent space, using a transformer to predict a denoising direction at each step. This formulation encompasses model families such as CogVideoX~\citep{yang2025cogvideox} and Wan~\citep{wan2025wan}. 
The sampling process takes place in a compressed spatio-temporal latent space. A video \(v\in\mathbb{R}^{T\times H\times W\times 3}\) corresponds to a latent \(z \in \mathbb{R}^{T_z\times H_z\times W_z\times C_z}\), where \(T_z\) and \((H_z, W_z)\) denote the latent temporal and spatial resolutions respectively, and \(C_z\) denotes the number of latent channels. A 3D Variational Autoencoder (VAE) is trained to decode the video \(v = D(z)\). Given a text prompt \(p\), a text encoder produces a sequence of tokens \(e_p\). Sampling begins with a Gaussian latent \(z_N\sim\mathcal{N}(0,I)\) and proceeds through \(N\) steps. At step \(k\), the Transformer \(\mathcal{T}_\theta\) predicts the update direction \(d_k\) as:
\[
 d_k=T_\theta(z_k,e_p,k),\qquad
 z_{k-1}=\operatorname{Step}_k(z_k,d_k).
\]
We refer to \(d_k\) as the update direction or \textit{denoising direction} interchangeably. 
\(\operatorname{Step}_k\) incorporates the sampling rule and update schedule, and can vary across models. The latent generated from the final step (\(z_0\)) is then de-compressed by the VAE decoder \(D\) to the output video \(v = D(z_0)\).

\paragraph{Classifier-free Guidance.}
Conditioning on the text prompt alone may not provide sufficient control over the content of the generated video. Classifier-free guidance (CFG)~\citep{ho2022cfg} was proposed to strengthen this conditioning during sampling, without requiring an external classifier. At each sampling step, CFG combines two update directions using the transformer:
\[
 d_{k,u}=T_\theta(z_k,e_{\mathrm{ref}},k),\qquad
 d_{k,c}=T_\theta(z_k,e_p,k),\qquad
 d_{k,\mathrm{cfg}}=d_{k,u}
   +\lambda_{\mathrm{cfg}}(d_{k,c}-d_{k,u}).
\]
Here, \(e_{\mathrm{ref}}\) represents the embedding of an empty prompt or a negative prompt. The update direction for step \(k\), \(d_{k,\mathrm{cfg}}\) is computed using the predictions from the unconditional or reference branch \(d_{k,u}\) and the prompt conditional branch \(d_{k,c}\). 
The strength of the extrapolation is controlled by the \textit{CFG guidance scale} \(\lambda_{\mathrm{cfg}} \in \mathbb{R}\), with different recommendations across models.

\section{TSG: Specification Guidance for Text-to-Video Generation}
\label{sec:tsg}

\newcommand{\TSGFrameDir}{ICLR_2027/images/tsg_method_frames}

\begin{figure*}[t]
    \centering
    \resizebox{\textwidth}{!}{%
      \input{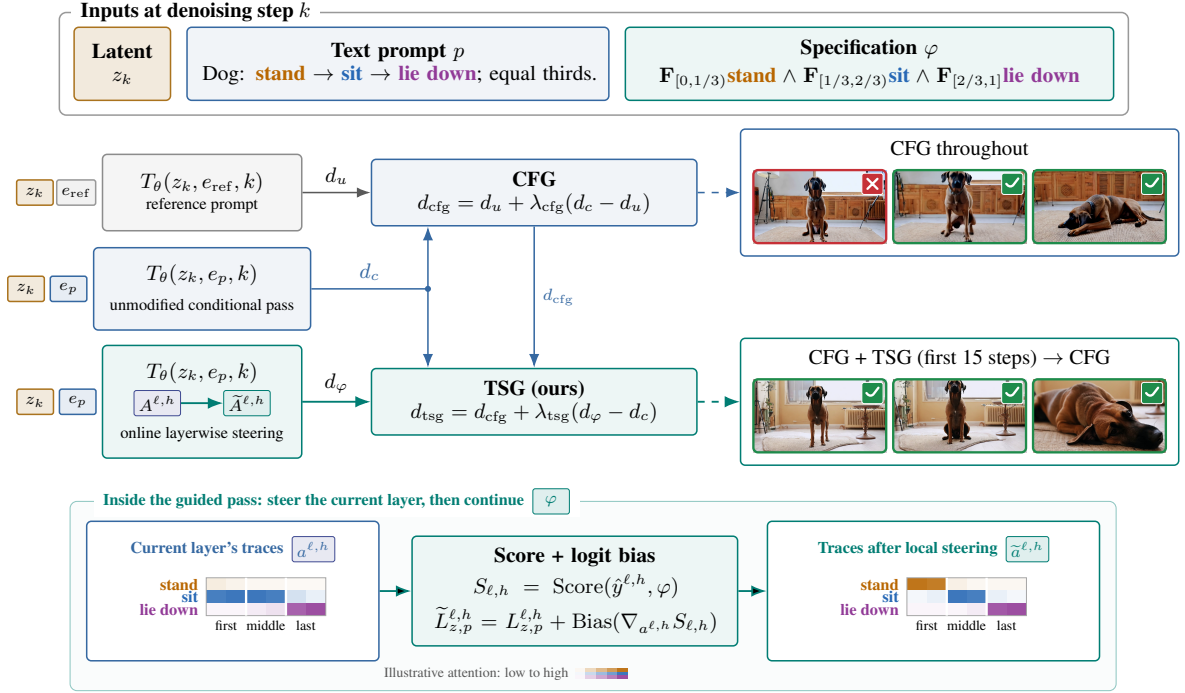}%
    }
    \caption{\emph{Temporal Specification Guidance (TSG).} During an additional conditional pass, TSG scores each head’s attention traces against the specification and applies gradient-derived logit biases. The resulting prediction augments CFG. Transformer weights receive no gradients. The example presents videos generated with Wan2.2 5B.}
    \label{fig:TSG}
    \vspace{-0.15in}
  \end{figure*}

We propose Temporal Specification Guidance (TSG), a training-free sampling-time framework that augments CFG with an update direction \(d_{k, \mathrm{tsg}}\) guided by the temporal specification \(\varphi\). 
This framework has three components: temporal traces that provide evidence of visual concepts, a differentiable scoring function that evaluates these traces against the specification \(\varphi\), and an update mechanism that uses gradients of this score to steer generation. In this work, we compute traces of visual concept evidence using video-text attention within the transformer \(T_\theta\), score them against \(\varphi\) with a differentiable quantitative semantics for \(MTL_f\), and use the gradients as attention logit biases.
This attention-steered forward pass produces the specification-guided direction \(d_{k,\varphi}\). We present the overall process in Figure~\ref{fig:TSG} and discuss each step in detail now.

\paragraph{Temporal Specification-guided sampling.}
Let \(d_{k,u}\) and \(d_{k,c}\) denote the reference and prompt-conditional direction predictions from Section~\ref{sec:background}. Similarly, let \(d_{k,\mathrm{cfg}}\) and \(\lambda_{k,\mathrm{cfg}}\) denote the CFG-predicted direction and the CFG guidance scale, respectively. At sampling step \(k\), TSG computes a \emph{specification-guided} direction \(d_{k,\varphi}\) using \(\varphi\) and yields the update direction \(d_{k,\mathrm{tsg}}\) as:
\[
 d_{k,\mathrm{tsg}}=d_{k,\mathrm{cfg}}
   +\lambda_{k,\mathrm{tsg}}(d_{k,\varphi}-d_{k,c}),
 \qquad
 z_{k-1}=\operatorname{Step}_k(z_k,d_{k,\mathrm{tsg}}).
\]

As with CFG, the strength of TSG is controlled by the \textit{TSG guidance scale} \(\lambda_{k,\mathrm{tsg}} \ge 0\). As the CFG guidance scale \(\lambda_{k,\mathrm{cfg}}\) usually takes values recommended by model providers and can dynamically vary across sampling steps, we compute \(\lambda_{\mathrm{tsg}}\) using the ratio between the TSG contribution and the \(\lambda_{\mathrm{cfg}}\)-scaled CFG contribution as:
\begin{equation}
\label{eq:tsg-scale}
\lambda_{k,\mathrm{tsg}}
=
\rho_{k,\mathrm{tsg}} \cdot\,
\frac{
  \textcolor{black!60}{
    \overbrace{
      \textcolor{black}{
        \left\| \lambda_{k,\mathrm{cfg}}(d_{k,\mathrm{c}}-d_{k,u})\right\|_2
      }
    }^{\text{Scaled CFG contribution}}
  }
}{
  \textcolor{black!60}{
    \underbrace{
      \textcolor{black}{
        \left\|d_{k,\varphi}-d_{k,c}\right\|_2
      }
    }_{\text{Unscaled TSG contribution}}
  }
}.
\end{equation}
The parameter \(\rho_{k,\mathrm{tsg}}\) denotes the relative strength of TSG with respect to CFG.
With \(\rho_{k,\mathrm{tsg}} = 1\), TSG's contribution is scaled to match that of CFG. 
We apply TSG for the first \(N_{\mathrm{tsg}}\) update steps, with CFG applied during the remaining steps. We next describe how the direction \(d_{k,\varphi}\) is computed.

\paragraph{Attention-derived temporal traces for visual concepts.}
To mitigate the need for an external concept classifier and inspired by recent work~\citep{schiber2026tempocontrol}, we use video-text attention to compute temporal traces of visual concepts. 
The text prompt \(p\) and video latent \(z\) provide tokens to serve as text keys and video queries, respectively, for every attention head \(h\) in layer \(\ell\) of transformer \(\mathcal{T}_\theta\). We map every concept proposition \(c\) in specification \(\varphi\) to the set of token indices \(P_c\) in prompt \(p\). Let \(Q_t\) be the set of spatial video queries for latent frame \(z_t\) and \(A[q,j]\) represent the post-softmax attention score for video token \(q \in Q_t\) and text token index \(j\). We construct the raw attention trace for concept \(c\) at layer \(\ell\) and head \(h\), \(a_{t,c}\) and the corresponding concept score trace, \(\hat{y}(z_t, c)\) as:
\begin{equation}
\label{eq:TSG-trace}
 a_{t,c}=\frac{1}{|\mathcal Q_t|}
 \sum_{q\in\mathcal Q_t}\max_{j\in\mathcal P_c} A[q,j],\qquad
 \hat y({z_t,c})=\sigma\!\left(\frac{a_{t,c}-\mu_c}{\tau s_c+\epsilon}\right).
\end{equation}
The score \(a_{t,c}\) corresponds to the maximum attention to any token corresponding to concept \(c\), averaged over all spatial positions for latent frame \(z_t\). We compute concept presence scores \(\hat{y}(z_t,c)\) by scaling attention scores over latent frames using the \(\operatorname{sigmoid}\) function with temporal mean \(\mu_c\) and population standard deviation \(s_c\). \(\tau>0\) is a temperature, and \(\epsilon>0\) stabilizes the denominator. 

Note that the latent frame \(z_t\) does not correspond to video frame \(v_t\) because it is defined at the latent temporal resolution \(T_z < T\). We limit steering to this latent resolution and convert normalized specification intervals to offsets on this grid using the convention in Section~\ref{sec:problem-setup}, with \(T\) replaced by \(T_z\). 

\paragraph{Differentiable Specification Scores with LogSTOP.}
Inspired by efficient quantitative semantics for Linear Temporal Logic specifications over videos~\citep{logstop}, we define an efficient interval-aware sum--product quantitative semantics in log space for \(MTL_f\) specifications. 
Given an MTL specification \(\varphi\) and concept score traces \(\hat y(z_t, c)\) at attention layer \(\ell\) and head \(h\), we can compute the score \(\hat y(z, \varphi)\) recursively using Algorithm~\ref{alg:logstop-mtl} in Appendix~\ref{appendix:logstop}.
Intuitively, conjunction and Always multiply support for their requirements, while disjunction, Eventually, and Until sum support for admissible alternatives or temporal witnesses.

The score for any specification \(\varphi\) can be computed using \(O(|\varphi|T_z)\) arithmetic operations (Proposition~\ref{prop:score-linear-time}).
The score is non-decreasing in independent literal-support inputs, as proved in Proposition~\ref{prop:score-monotonicity}. This yields predictable gradient signs for concept--time inputs with consistent polarity (Corollary~\ref{cor:score-polarity}). The direction is determined by the surrounding trace when a concept appears with both polarities. Appendix~\ref{appendix:logstop} provides further details on this monotonicity property, the algorithm's complexity, and examples of the resulting gradients.

\paragraph{Online attention steering.}
For each layer \(\ell\), head \(h\), and concept \(c \in \varphi\), we differentiate \(\hat{y}(z_t,\varphi)\) with respect to the raw trace \(a_{t,c}\):
\[
 g_{\cdot,c}=\nabla_{a_{\cdot,c}}\hat{y}(z_t, \varphi),\qquad
 b_{\cdot,c}=\eta\,g_{\cdot,c}/\|g_{\cdot,c}\|_\infty,\qquad
 B[q,j]=\sum_{c:\,j\in\mathcal P_c} b_{t(q),c}.
\]
where \(b_{t,c}\) is the normalized \textit{attention bias} for temporal position \(t\) and concept \(c\). We set \(b_{\cdot,c}=0\) for negligible gradients. The step size \(\eta\) controls the bias magnitude. We broadcast temporal bias \(b_{t,c}\) to every spatial query \(q \in Q_t\) at position \(t(q)\) and every text token \(j\) associated with the concept \(c\) as \(B[q,j]\).
For spatial query \(q\), we further center the biases by the mean bias over all text keys and \(\operatorname{clip}\) them in a given range \([b_{\min}, b_{\max}]\). 

We apply the resulting bias to the logits for layer \(\ell\) and head \(h\) as:
\begin{equation}
\label{eq:TSG-attention-bias}
\widetilde L_{z,p}^{\ell,h}=L_{z,p}^{\ell,h}+B^{\ell,h}
\end{equation}
The biased attention is used to compute the layer's output, and the forward pass continues with the updated representations. Biases are computed anew at each selected layer and sampling step, separately for each head. The completed guided pass yields the specification-guided direction \(d_{k,\varphi}\).

\section{Experimental Evaluation}
\label{sec:experiments}

% Pictorial summaries of Tsg v4 examples (normalized video time).
% Source IDs: 2st-181, tst-001, rsr-081, oe2-201, oe3-121, hwo-021,
% exr-061, pst-101, trg-041, aec-141, afs-161.
\begin{figure}[t]
\centering
\begingroup
\definecolor{benchblue}{RGB}{45,83,117}
\definecolor{benchgreen}{RGB}{45,105,89}
\definecolor{benchpurple}{RGB}{104,76,126}
% Filled vector silhouettes retain distinct poses at publication size.
\newcommand{\bpstand}{%
 \fill (0,.95) circle (.12);
 \fill[rounded corners=1pt] (-.13,.78)--(.13,.78)--(.2,.48)--(.13,.45)
 --(.1,.63)--(.1,.4)--(.15,.03)--(.02,.03)--(0,.36)
 --(-.04,.03)--(-.17,.03)--(-.1,.4)--(-.1,.63)--(-.15,.44)
 --(-.23,.48)--(-.17,.71)--cycle;
 \fill (-.18,0) rectangle (-.04,.07);\fill (.02,0) rectangle (.18,.07);}
\newcommand{\bpsit}{%
 \fill (-.22,.73) circle (.12);
 \fill (-.31,.55) .. controls (-.17,.63) and (-.1,.51) .. (-.13,.37)
 --(-.19,.09) .. controls (-.2,0) and (-.35,0) .. (-.36,.07)
 --(-.34,.19)--cycle;
 % Hips and heels share the floor baseline.
 \fill (-.34,.13)--(.06,.35)--(.3,.11)--(.45,.11)
 arc (90:-90:.055)--(.24,0)--(.02,.21)--(-.23,0)--(-.35,0)--cycle;
 \fill (-.21,.5)--(-.13,.53)--(.02,.3)--(.16,.26)--(.14,.18)--(-.05,.22)--cycle;
 \draw[black!40,line width=.4pt] (-.43,0)--(.51,0);}
% Same muzzle and floppy ear in all three dog poses.
\newcommand{\bdhead}[2]{\begin{scope}[shift={(#1,#2)}]
 \fill (-.13,-.09) .. controls (-.23,.14) and (-.08,.26) .. (.09,.19)
 --(.17,.11)--(.34,.06)--(.33,-.04)--(.17,-.1)--(.04,-.1)--cycle;
 \fill[black!90] (-.11,.17) .. controls (-.3,.18) and (-.26,-.14) .. (-.12,-.13)
 .. controls (-.03,-.09) and (-.07,.08) .. (-.11,.17);
 \fill[white] (.09,.1) circle (.022);
 \fill[black!90] (.32,.055) circle (.035);
 \end{scope}}
\newcommand{\bdog}{%
 \fill (-.08,.4) ellipse (.37 and .155);
 \fill (-.4,.41)--(-.43,.05)--(-.29,.05)--(-.28,.33)
 --(-.1,.33)--(-.08,.05)--(.04,.05)--(0,.34)--cycle;
 \fill (.15,.46)--(.18,.05)--(.32,.05)--(.3,.44)--(.37,.56)--(.18,.6)--cycle;
 \draw[line width=2.3pt] (-.37,.44) .. controls (-.62,.51) and (-.57,.66) .. (-.6,.68);
 \bdhead{.3}{.62}}
\newcommand{\bdsit}{%
 \fill (-.22,.22) ellipse (.24 and .19);
 \fill (-.35,.25) .. controls (-.21,.5) and (-.15,.55) .. (.04,.62)
 --(.19,.45)--(.11,.2)--(-.12,.09)--cycle;
 \fill (.06,.46)--(.08,.04)--(.32,.04)--(.32,.11)--(.22,.14)--(.23,.49)--cycle;
 \fill (-.36,.04) rectangle (-.07,.13);
 \draw[line width=2.2pt] (-.3,.13) .. controls (-.57,.02) and (-.66,.14) .. (-.56,.23);
 \bdhead{.14}{.62}}
\newcommand{\bdlie}{%
 \fill (-.13,.2) ellipse (.39 and .17);
 \fill (-.21,.04) rectangle (.56,.13);
 \fill (.14,.14)--(.19,.32)--(.4,.34)--(.4,.14)--cycle;
 \draw[line width=2pt] (-.42,.15) .. controls (-.65,.1) and (-.62,.3) .. (-.54,.32);
 \bdhead{.25}{.35}}
\newcommand{\bcat}{%
 \fill (-.06,.32) ellipse (.3 and .13);
 \fill (.15,.31)--(.12,.66)--(.26,.55)--(.39,.67)--(.42,.37)--(.32,.27)--cycle;
 \foreach \x in {-.3,-.12,.17,.32}{\fill (\x,.31) rectangle ++(.07,-.29);}
 \draw[line width=1.9pt] (-.33,.35) .. controls (-.67,.55) and (-.35,.93) .. (-.43,.67);
 \fill[white] (.29,.49) circle (.023);}
\newcommand{\brabbit}{%
 \fill (-.08,.23) ellipse (.28 and .21);
 \fill (.2,.35) ellipse (.19 and .15);
 \fill[rotate around={12:(.12,.57)}] (.12,.57) ellipse (.065 and .3);
 \fill[rotate around={-10:(.27,.6)}] (.27,.6) ellipse (.06 and .3);
 \fill (-.36,.25) circle (.1);
 \fill (-.12,.02) rectangle (.24,.09);
 \fill[white] (.28,.39) circle (.024);}
\newcommand{\bhorse}{%
 \fill (-.12,.51) ellipse (.35 and .19);
 \fill (.07,.53)--(.18,.95)--(.33,1.04)--(.49,.94)--(.63,.75)
 --(.53,.67)--(.36,.77)--(.29,.44)--cycle;
 \fill (.24,.98)--(.22,1.12)--(.33,1.03)--cycle;
 \foreach \x in {-.4,-.22,.1,.24}{\fill (\x,.49)--(\x+.08,.49)--(\x+.11,.04)--(\x-.02,.04)--cycle;}
 \fill (-.4,.61) .. controls (-.66,.6) and (-.51,.22) .. (-.68,.14)
 .. controls (-.43,.17) and (-.44,.41) .. (-.4,.61);
 \fill[white] (.39,.91) circle (.023);}
\newcommand{\btree}{%
 \fill (-.045,0) rectangle (.045,.65);
 \fill (-.28,.45) .. controls (-.6,.6) and (-.4,.92) .. (-.17,.9)
 .. controls (-.13,1.2) and (.24,1.17) .. (.3,.94)
 .. controls (.64,.9) and (.57,.5) .. (.3,.48)--cycle;}
\newcommand{\bicon}[4]{\begin{scope}[shift={(#2,#3)},scale=#4,
 draw=black!75,fill=black!75,line width=1.1pt,line cap=round,line join=round]
 \csname b#1\endcsname\end{scope}}
\newcommand{\blight}[3]{%
 \draw[rounded corners=2pt,fill=black!75,draw=none] (#1-.18,-1.85) rectangle (#1+.18,-.93);
 \foreach \yy in {-1.1,-1.39,-1.68}{\fill[white!35!black] (#1,\yy) circle (.105);}
 \fill[#3] (#1,#2) circle (.105);}
% Panel content uses local coordinates. All times are fractions of duration.
\newcommand{\bpanel}[6]{\begin{scope}[shift={(#1,#2)}]
 \draw[draw=black!18,rounded corners=2pt,fill=white] (0,0) rectangle (4.48,-2.08);
 \node[anchor=north west,text width=4.18cm,align=left,inner sep=0pt,
 font=\fontsize{10}{11}\selectfont\bfseries,text=#3] at (.15,-.13) {#4};
 \begin{scope}[yshift=.30cm]
 \fill[#3!7] (.15,-.87) rectangle (4.33,-1.98);
 \draw[black!45,line width=.4pt] (.15,-1.98)--(4.33,-1.98);
 \ifnum#5=2
  \draw[white,line width=1pt] (2.24,-.87)--(2.24,-1.98);
  \foreach \x/\t in {.15/0,2.24/{1/2},4.33/1}{
   \draw[black!45] (\x,-1.98)--(\x,-2.04);
   \node[font=\fontsize{8}{9}\selectfont,anchor=north,inner sep=1pt] at (\x,-2.05) {$\t$};}
 \else
  \foreach \x in {1.543,2.937}{\draw[white,line width=1pt] (\x,-.87)--(\x,-1.98);}
  \ifnum#5=4
   \def\bticks{.15/0,1.543/{1/3},4.33/1}
  \else
   \def\bticks{.15/0,1.543/{1/3},2.937/{2/3},4.33/1}
  \fi
  \foreach \x/\t in \bticks{
   \draw[black!45] (\x,-1.98)--(\x,-2.04);
   \node[font=\fontsize{8}{9}\selectfont,anchor=north,inner sep=1pt] at (\x,-2.05) {$\t$};}
 \fi
 #6\end{scope}\end{scope}}
\resizebox{\linewidth}{!}{\begin{tikzpicture}
\bpanel{0}{0}{benchblue}{Two-state transition}{2}{
 \bicon{pstand}{1.2}{-1.87}{.85}\bicon{psit}{3.28}{-1.87}{.85}}
\bpanel{4.7}{0}{benchblue}{Three-state transition}{3}{
 \blight{.85}{-1.39}{yellow!85!orange}\blight{2.24}{-1.1}{red!75!black}\blight{3.63}{-1.68}{green!65!black}}
\bpanel{9.4}{0}{benchblue}{Reversible state return}{3}{
 \bicon{pstand}{.85}{-1.87}{.85}\bicon{psit}{2.24}{-1.87}{.85}\bicon{pstand}{3.63}{-1.87}{.85}}
\bpanel{14.1}{0}{benchblue}{Persistent-entity transition}{2}{
 \bicon{pstand}{.75}{-1.87}{.85}\bicon{dog}{1.62}{-1.87}{.7}
 \bicon{pstand}{2.85}{-1.87}{.85}\bicon{dsit}{3.72}{-1.87}{.7}}
\bpanel{0}{-2.28}{benchgreen}{Ordered entry (one)}{2}{
 \bicon{pstand}{1.2}{-1.87}{.85}\bicon{pstand}{2.86}{-1.87}{.85}\bicon{dog}{3.7}{-1.87}{.65}}
\bpanel{4.7}{-2.28}{benchgreen}{Ordered entry (two)}{3}{
 \bicon{dog}{.85}{-1.75}{.65}
 \bicon{dog}{1.95}{-1.42}{.5}\bicon{cat}{2.52}{-1.86}{.5}
 \bicon{dog}{3.3}{-1.4}{.48}\bicon{cat}{3.91}{-1.48}{.48}\bicon{rabbit}{3.58}{-1.91}{.48}}
\bpanel{9.4}{-2.28}{benchgreen}{Overlapping handover}{3}{
 \bicon{horse}{.85}{-1.86}{.75}
 \bicon{horse}{1.99}{-1.85}{.75}\bicon{rabbit}{2.62}{-1.86}{.47}
 \bicon{rabbit}{3.63}{-1.86}{.75}}
\bpanel{14.1}{-2.28}{benchgreen}{Entity exit and re-entry}{3}{
 \bicon{dog}{.85}{-1.85}{.8}\node[text=black!45] at (2.24,-1.45) {$\emptyset$};\bicon{dog}{3.63}{-1.85}{.8}}
\bpanel{0}{-4.56}{benchpurple}{Triggered transition}{3}{
 \bicon{dsit}{.85}{-1.87}{.62}
 \bicon{pstand}{1.89}{-1.87}{.85}\bicon{dsit}{2.58}{-1.87}{.62}
 \bicon{pstand}{3.25}{-1.87}{.85}\bicon{dog}{3.9}{-1.87}{.62}}
% The two right-hand scenes are alternatives over one shared time interval.
\bpanel{4.7}{-4.56}{benchpurple}{Anchored entity choice}{4}{
 \bicon{tree}{.85}{-1.88}{.75}
 \bicon{tree}{2.1}{-1.88}{.75}\bicon{dog}{2.5}{-1.88}{.42}
 \bicon{tree}{3.49}{-1.88}{.75}\bicon{cat}{3.92}{-1.88}{.42}
 \node[font=\fontsize{8}{9}\selectfont\bfseries,fill=white,inner sep=2pt]
  at (2.937,-1.44) {or};}
\bpanel{9.4}{-4.56}{benchpurple}{Alternative final state}{4}{
 \bicon{dsit}{.85}{-1.87}{.8}
 \bicon{dog}{2.24}{-1.87}{.8}\bicon{dlie}{3.63}{-1.87}{.8}
 \node[font=\fontsize{8}{9}\selectfont\bfseries,fill=white,inner sep=2pt]
  at (2.937,-1.44) {or};}
% Uniform labels: entity names above, entity and pose below.
\begin{scope}[shift={(14.1,-4.56)},
 every node/.style={font=\fontsize{8}{9}\selectfont,inner sep=0pt,text=black!75,align=center}]
 \bicon{cat}{.6}{-.51}{.43}
 \bicon{horse}{1.68}{-.51}{.38}
 \bicon{rabbit}{2.75}{-.51}{.43}
 \bicon{tree}{3.85}{-.51}{.38}
 \foreach \x/\label in {.6/cat,1.68/horse,2.75/rabbit,3.85/tree}{
  \node at (\x,-.69) {\label};}
 \bicon{pstand}{.42}{-1.42}{.45}
 \bicon{psit}{1.29}{-1.42}{.45}
 \bicon{dog}{2.2}{-1.42}{.45}
 \bicon{dsit}{3.1}{-1.42}{.45}
 \bicon{dlie}{4}{-1.42}{.45}
 \foreach \x/\entity/\pose in {.42/person/stands,1.29/person/sits,2.2/dog/stands,3.1/dog/sits,4/dog/lies}{
  \node[anchor=north] at (\x,-1.58) {\strut\entity\\\strut\pose};}
\end{scope}
\end{tikzpicture}}
\endgroup
\caption{\textbf{TSG benchmark with interval-timed temporal requirements from 11 categories.} Shaded intervals indicate permitted event timing within the video duration; $\emptyset$ denotes absence, and ``or'' denotes mutually exclusive outcomes. Pictograms are illustrative.}
\vspace{-0.2in}
\label{fig:benchmark-categories}
\end{figure}
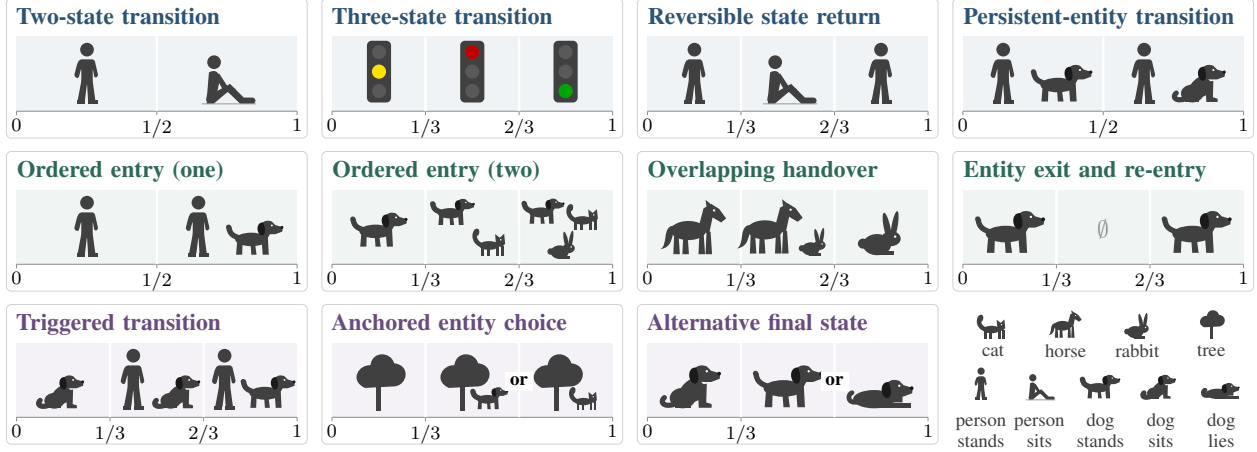

\subsection{Setup and Benchmarks}

\paragraph{The TSG Benchmark with interval-timed temporal constraints.}
Existing benchmarks primarily focus on untimed constraints, such as state transitions in TC-Bench~\citep{tcbench} and NeuS-V~\citep{neusv}, and strictly-timed constraints such as  TempoControl~\citep{schiber2026tempocontrol}. To evaluate TSG on interval-timed constraints, we construct a benchmark with 220 text prompts from 11 categories of temporal constraints. These categories are presented in Figure~\ref{fig:benchmark-categories} and include variations in the requested timing, ordering, persistence, and choice among visual concepts.  

We use \texttt{GPT-5.6-Sol} to propose 50 entities spanning objects from VBench~\citep{vbench} and background settings, with associated states (e.g., pose of an object, or time of day for a setting). We construct sets of 20 prompts for each temporal constraint category in Figure~\ref{fig:benchmark-categories}. We restrict the intervals to halves and thirds within 5--6-second videos, allowing transitions to occur at any point within each interval. Finally, we manually assign temporal specification templates for each category and instruct \texttt{GPT-5.6-Sol} to instantiate the specifications with atomic propositions mapped to respective visual concepts in each prompt. Appendix~\ref{appendix:benchmark} provides further details on examples of prompts for each category along with the list of temporal specification templates.

\paragraph{Baselines and Models.}
We evaluate text-to-video generation models with CFG sampling as our main baseline. We include five models from two model families: CogVideoX-2B and CogVideoX-5B~\citep{yang2025cogvideox}, and Wan2.1 T2V-1.3B, Wan2.2 TI2V-5B, and Wan2.2 T2V A14B~\citep{wan2025wan}. These models cover both joint video--text attention, used by CogVideoX, and dedicated video-to-text cross-attention, used by Wan. We use the CFG settings recommended by the model providers.
On the strictly-timed prompts from the TempoControl benchmark, we also compare TSG against the TempoControl method. 
We use TSG during the first 15 sampling steps with \(\rho_{k,\mathrm{tsg}}=1\).
Appendices~\ref{appendix:models} and~\ref{appendix:tsg-settings} describe the model settings and TSG configuration; Appendices~\ref{appendix:tempocontrol} and~\ref{appendix:compute} detail the TempoControl comparison and compute measurements.

\begin{table*}[t]
\caption{\textbf{TSG improves satisfaction accuracy on a benchmark of 220 prompts.} Latency and Memory measurements are on B200 for all models, except for Wan2.2 A14B which was run on H200.}
\centering
\small
\setlength{\tabcolsep}{5.7pt}
\begin{tabular}{llHHlHHlll}
\toprule
Model & Method & Config & (stale) Timed accuracy (\%) & Sat Acc. (\%) & Coverage & Prop. Recall (Micro) & Img. quality & Latency (min) & Peak GPU mem. \\
\midrule
CogVideoX 2B & Base & -- & 8.2\% \textcolor{gray}{(+0.0\%)} & 10.5\% \textcolor{gray}{(+0.0\%)} & 85.70\% \textcolor{gray}{(+0.0\%)} & 73.80\% \textcolor{gray}{(+0.0\%)} & 59.80 \textcolor{gray}{(+0.0)} & 0.75 $\pm$ 0.03 \textcolor{gray}{($1.00\times$)} & 26.30 GiB \\
 & TSG & rho=1, step=1, layers=0:1, guidance=15 & 15.5\% \textcolor{green!50!black}{(+7.3\%)} & 30.5\% \textcolor{green!50!black}{(+20.0\%)} & 89.66\% \textcolor{green!50!black}{(+4.0\%)} & 86.20\% \textcolor{green!50!black}{(+12.4\%)} & 56.85 \textcolor{red}{(-3.0)} & 2.62 $\pm$ 0.38 \textcolor{red}{($3.49\times$)} & 26.33 GiB \\
\midrule
CogVideoX 5B & Base & -- & 6.8\% \textcolor{gray}{(+0.0\%)} & 8.6\% \textcolor{gray}{(+0.0\%)} & 86.16\% \textcolor{gray}{(+0.0\%)} & 78.40\% \textcolor{gray}{(+0.0\%)} & 63.72 \textcolor{gray}{(+0.0)} & 1.82 $\pm$ 0.08 \textcolor{gray}{($1.00\times$)} & 31.28 GiB \\
 & TSG & rho=1, step=1, layers=0:1, guidance=15 & 17.3\% \textcolor{green!50!black}{(+10.5\%)} & 26.8\% \textcolor{green!50!black}{(+18.2\%)} & 90.76\% \textcolor{green!50!black}{(+4.6\%)} & 85.20\% \textcolor{green!50!black}{(+6.8\%)} & 62.18 \textcolor{red}{(-1.5)} & 6.06 $\pm$ 0.86 \textcolor{red}{($3.33\times$)} & 31.31 GiB \\
\midrule
Wan 2.1 1.3B & Base & -- & 11.8\% \textcolor{gray}{(+0.0\%)} & 15.9\% \textcolor{gray}{(+0.0\%)} & 85.57\% \textcolor{gray}{(+0.0\%)} & 87.80\% \textcolor{gray}{(+0.0\%)} & 69.25 \textcolor{gray}{(+0.0)} & 1.39 $\pm$ 0.02 \textcolor{gray}{($1.00\times$)} & 30.48 GiB \\
 & TSG & rho=1, step=1, layers=0:1, guidance=15 & 20.5\% \textcolor{green!50!black}{(+8.6\%)} & 31.4\% \textcolor{green!50!black}{(+15.5\%)} & 89.07\% \textcolor{green!50!black}{(+3.5\%)} & 93.40\% \textcolor{green!50!black}{(+5.6\%)} & 67.39 \textcolor{red}{(-1.9)} & 2.35 $\pm$ 0.17 \textcolor{red}{($1.69\times$)} & 30.52 GiB \\

\midrule

Wan 2.2 5B & Base & -- & 15.5\% \textcolor{gray}{(+0.0\%)} & 25.0\% \textcolor{gray}{(+0.0\%)} & 85.46\% \textcolor{gray}{(+0.0\%)} & 88.80\% \textcolor{gray}{(+0.0\%)} & 70.38 \textcolor{gray}{(+0.0)} & 2.43 $\pm$ 0.03 \textcolor{gray}{($1.00\times$)} & 46.83 GiB \\
 & TSG & rho=1, step=1, layers=0:1, guidance=15 & 24.5\% \textcolor{green!50!black}{(+9.1\%)} & 34.5\% \textcolor{green!50!black}{(+9.5\%)} & 88.02\% \textcolor{green!50!black}{(+2.6\%)} & 90.80\% \textcolor{green!50!black}{(+2.0\%)} & 69.16 \textcolor{red}{(-1.2)} & 4.56 $\pm$ 0.41 \textcolor{red}{($1.87\times$)} & 45.15 GiB \\

\midrule

Wan 2.2 A14B & Base & -- & 10.9\% \textcolor{gray}{(+0.0\%)} & 35.9\% \textcolor{gray}{(+0.0\%)} & 85.71\% \textcolor{gray}{(+0.0\%)} & 92.80\% \textcolor{gray}{(+0.0\%)} & 65.42 \textcolor{gray}{(+0.0)} & 24.28 $\pm$ 0.21 \textcolor{gray}{($1.00\times$)} & 109.33 GiB \\
 & TSG & rho=1, step=1, layers=0:1, guidance=15 & 21.4\% \textcolor{green!50!black}{(+10.5\%)} & 46.8\% \textcolor{green!50!black}{(+10.9\%)} & 89.34\% \textcolor{green!50!black}{(+3.6\%)} & 95.60\% \textcolor{green!50!black}{(+2.8\%)} & 68.63 \textcolor{green!50!black}{(+3.2)} & 35.18 $\pm$ 1.93 \textcolor{red}{($1.45\times$)} & 107.95 GiB \\

\bottomrule
\end{tabular}
\label{tab:TSG-results}
\vspace{-0.1in}
\end{table*}

\paragraph{Metrics.}
We uniformly sample 21 decoded frames from each video, and evaluate the frame-level presence of a visual concept using \texttt{Qwen2.5-VL-7B-Instruct}.
% \textcolor{gray}{TODO: We present the prompts used with this model, along with correlation with human judgements in Appendix.} 
We use the ground-truth semantics of MTL described in Appendix~\ref{appendix:mtl_semantics} to compute the specification satisfaction accuracy (Sat Acc.). We also evaluate visual quality of the generated videos using VBench's MUSIQ-SPAQ imaging quality scorer~\citep{vbench}, along with the time taken for generation (latency) and the peak GPU memory allocated to the process. 
Appendix~\ref{appendix:metrics} provides further details.

\subsection{RQ1. Does TSG improve adherence to temporal specifications?}
\label{sec:experiments-tsg}

\paragraph{Result 1: TSG consistently outperforms CFG-only on interval-timed and strictly-timed temporal constraints.} 
Table~\ref{tab:TSG-results} presents the results comparing TSG against CFG-only sampling on the TSG benchmark. TSG improves satisfaction accuracy by 9.5--20 percentage points across models, with Wan 2.2 A14B reporting the highest satisfaction accuracy of \(46.8\%\) with TSG. Interestingly,
Wan2.1 1.3B reports a \(31.4\%\) accuracy with TSG compared to Wan2.2 5B's \(25\%\) with CFG-only sampling, despite having fewer parameters. We see this pattern more broadly: \textit{smaller models with TSG report higher or comparable satisfaction accuracies to larger models with CFG only}. On the strictly-timed prompts from the TempoControl benchmark, TSG improves temporal accuracy over CFG-only by 17.6--22.0 percentage points across splits, and exceeds full TempoControl by 2.97 and 0.58 percentage points on the one-object and two-object splits, respectively (Table~\ref{tab:tempocontrol-results}).

\subsection{RQ2. What are the computational and quality costs of TSG?}

\paragraph{Result 2: TSG incurs little to no memory overhead but increases time taken for generation.}
TSG matches the CFG-only sampling in terms of the peak GPU memory allocated to the model. This is expected as attention steering detaches the computational graph and hence does not need to store gradients with respect to model weights. TSG uses 30.52\,GiB versus 79.43\,GiB for full TempoControl, a reduction of approximately 62\%, as TempoControl optimizes video latents requiring a gradient computation graph through the entire video transformer (Table~\ref{tab:tempocontrol-results}).

The additional specification-guided forward pass through the transformer incurs a latency cost with TSG. This scales with the size of the model as expected, with TSG taking around 1 extra minute with Wan2.1 1.3B to 11 extra minutes with Wan2.2 A14B (Table~\ref{tab:TSG-results}). Full TempoControl takes approximately $4.9\times$ and $3.6\times$ as long as TSG on the one-object and two-object splits, respectively (Table~\ref{tab:tempocontrol-results}). Limiting TempoControl to one optimization round brings its latency close to TSG's but retains approximately 79.4\,GiB peak memory; TSG improves temporal accuracy over this variant by 20.1\% and 27.5\% relative to its one-object and two-object scores.

\paragraph{Result 3: TSG modestly impacts the visual quality of the generated videos.}
TSG lowers imaging quality by up to \(3.0\) points on four models and improves it by \(3.2\) points on Wan2.2 A14B (Table~\ref{tab:TSG-results}).

% Requires \usepackage{booktabs,xcolor}
\begin{table*}[t]
\caption{
\textbf{TempoControl benchmark results with Wan2.1 1.3B.} Parentheses show accuracy changes in percentage points and computational-cost ratios relative to CFG. Latency reports mean ± standard deviation; memory reports peak allocation. We present setup details in Appendix~\ref{appendix:tempocontrol}.
}
\label{tab:tempocontrol-results}
\centering
\small
\begin{tabular}{HlHlHHll}
\toprule

Model & Method & Config & Temporal Acc $\uparrow$ & Absence Acc $\uparrow$ & Presence Acc $\uparrow$ & Latency (min/video) $\downarrow$ & Peak GPU mem (GiB) $\downarrow$ \\
\midrule
\multicolumn{8}{l}{\textbf{One object} (N=80)} \\
\midrule

Wan 2.1 T2V 1.3B & Base & -- & 69.40\% \textcolor{gray}{(0.0\%)} & 68.12\% \textcolor{gray}{(0.0\%)} & 70.57\% \textcolor{gray}{(0.0\%)} & 1.29 $\pm$ 0.01 \textcolor{gray}{(1.00$\times$)} & 30.48 \textcolor{gray}{(1.00$\times$)} \\

Wan 2.1 T2V 1.3B & TempoControl (one round) & -- & 72.44\% \textcolor{green!50!black}{(+3.0\%)} & 71.25\% \textcolor{green!50!black}{(+3.1\%)} & 73.52\% \textcolor{green!50!black}{(+3.0\%)} & 2.23 $\pm$ 0.01 \textcolor{red}{(1.72$\times$)} & 79.41 \textcolor{red}{(2.61$\times$)} \\

Wan 2.1 T2V 1.3B & TempoControl (full) & -- & 84.05\% \textcolor{green!50!black}{(+14.6\%)} & 84.62\% \textcolor{green!50!black}{(+16.5\%)} & 83.52\% \textcolor{green!50!black}{(+13.0\%)} & 9.87 $\pm$ 2.10 \textcolor{red}{(7.63$\times$)} & 79.43 \textcolor{red}{(2.61$\times$)} \\

Wan 2.1 T2V 1.3B & TSG & rho=1.0, layers=0:1 & 87.02\% \textcolor{green!50!black}{(+17.6\%)} & 91.12\% \textcolor{green!50!black}{(+23.0\%)} & 83.30\% \textcolor{green!50!black}{(+12.7\%)} & 2.02 $\pm$ 0.01 \textcolor{red}{(1.56$\times$)} & 30.52 \textcolor{gray}{(1.00$\times$)} \\
\midrule
\multicolumn{8}{l}{\textbf{Two object} (N=82)} \\
\midrule
Wan 2.1 T2V 1.3B & Base & -- & 36.12\% \textcolor{gray}{(0.0\%)} & 47.32\% \textcolor{gray}{(0.0\%)} & 25.94\% \textcolor{gray}{(0.0\%)} & 1.30 $\pm$ 0.01 \textcolor{gray}{(1.00$\times$)} & 30.48 \textcolor{gray}{(1.00$\times$)} \\

Wan 2.1 T2V 1.3B & TempoControl (one round) & -- & 45.53\% \textcolor{green!50!black}{(+9.4\%)} & 51.10\% \textcolor{green!50!black}{(+3.8\%)} & 40.47\% \textcolor{green!50!black}{(+14.5\%)} & 2.28 $\pm$ 0.10 \textcolor{red}{(1.75$\times$)} & 79.42 \textcolor{red}{(2.61$\times$)} \\

Wan 2.1 T2V 1.3B & TempoControl (full) & -- & 57.49\% \textcolor{green!50!black}{(+21.4\%)} & 63.90\% \textcolor{green!50!black}{(+16.6\%)} & 51.66\% \textcolor{green!50!black}{(+25.7\%)} & 8.11 $\pm$ 3.41 \textcolor{red}{(6.24$\times$)} & 79.43 \textcolor{red}{(2.61$\times$)} \\

Wan 2.1 T2V 1.3B & TSG & rho=1.0, layers=0:1 & 58.07\% \textcolor{green!50!black}{(+22.0\%)} & 70.00\% \textcolor{green!50!black}{(+22.7\%)} & 47.23\% \textcolor{green!50!black}{(+21.3\%)} & 2.27 $\pm$ 0.01 \textcolor{red}{(1.74$\times$)} & 30.52 \textcolor{gray}{(1.00$\times$)} \\

\bottomrule
\end{tabular}
\vspace{-0.1in}
\end{table*}

\section{Analysis and Discussion}
\label{sec:discussion}

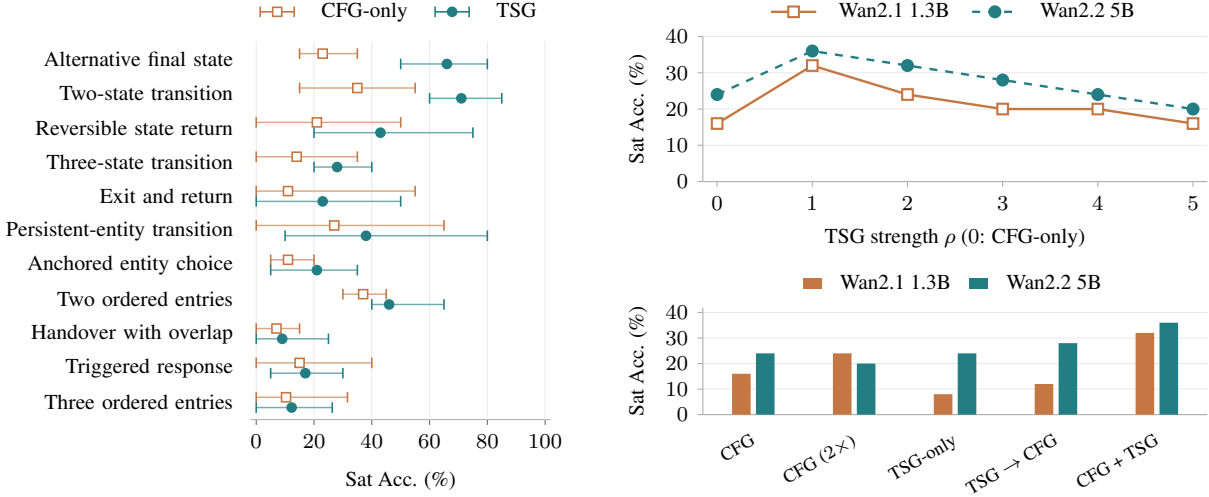
\begin{figure*}[t]
      \centering
      \begin{subfigure}[t]{0.48\textwidth}
          \centering
          \vspace{0pt}
          % Mean accuracy; asymmetric whiskers are min and max across the five models.
% Category order: descending mean TSG minus base gain.
% Requires pgfplots, compat=1.18. No title or footnotes.
\providecommand{\categoryplotfont}{\normalfont}
\begin{tikzpicture}
\definecolor{basecolor}{HTML}{C47743}
\definecolor{tsgcolor}{HTML}{237E83}
\definecolor{rangegrid}{HTML}{E9E9E9}
\definecolor{rangeaxis}{HTML}{666666}
\definecolor{rangeborder}{HTML}{B0B0B0}
\begin{axis}[
width=\dimexpr\linewidth-3.95cm\relax,height=5cm,scale only axis,
xmin=-2,xmax=102,ymin=-0.5,ymax=10.5,y dir=reverse,
xtick={0,20,40,60,80,100},ytick={0,1,2,3,4,5,6,7,8,9,10},
yticklabels={
{Alternative final state},
{Two-state transition},
{Reversible state return},
{Three-state transition},
{Exit and return},
{Persistent-entity transition},
{Anchored entity choice},
{Two ordered entries},
{Handover with overlap},
{Triggered response},
{Three ordered entries}
},font=\categoryplotfont\fontsize{8}{9}\selectfont,tick label style={font=\categoryplotfont\fontsize{8}{9}\selectfont},
yticklabel style={anchor=east,xshift=-3pt,text width=3.0cm,align=right},
xlabel={Sat Acc. (\%)},
xlabel style={align=center,font=\categoryplotfont\fontsize{8}{9}\selectfont,yshift=-2pt},
axis x line*=bottom,axis y line*=left,axis line style={rangeborder,line width=0.4pt},y axis line style={draw=none},yticklabel pos=left,xtick pos=bottom,ytick pos=left,
xtick align=outside,major tick length=2pt,tick style={rangeaxis,line width=0.4pt},ytick style={draw=none},
legend style={at={(0.5,1.025)},anchor=south,draw=none,fill=none,legend columns=2,cells={anchor=west},column sep=8pt,font=\categoryplotfont\fontsize{8}{9}\selectfont},clip=true,
]
\draw[rangegrid,line width=0.3pt] (axis cs:0,-0.5) -- (axis cs:0,10.5);
\draw[rangegrid,line width=0.3pt] (axis cs:20,-0.5) -- (axis cs:20,10.5);
\draw[rangegrid,line width=0.3pt] (axis cs:40,-0.5) -- (axis cs:40,10.5);
\draw[rangegrid,line width=0.3pt] (axis cs:60,-0.5) -- (axis cs:60,10.5);
\draw[rangegrid,line width=0.3pt] (axis cs:80,-0.5) -- (axis cs:80,10.5);
\draw[rangegrid,line width=0.3pt] (axis cs:100,-0.5) -- (axis cs:100,10.5);
\addplot+[forget plot,only marks,mark=square*,mark size=1.7pt,color=basecolor,
mark options={fill=white,draw=basecolor,line width=0.6pt},
error bars/.cd,x dir=both,x explicit,error mark=|,
error bar style={draw=basecolor!65,line width=0.55pt},
error mark options={rotate=0,draw=basecolor,mark size=2pt,line width=0.6pt}
] table[x=mean,y=row,x error minus=minus,x error plus=plus,row sep=\\] {
mean row minus plus\\
23.0000000000 -0.15 8.0000000000 12.0000000000\\
35.0000000000 0.85 20.0000000000 20.0000000000\\
21.0000000000 1.85 21.0000000000 29.0000000000\\
14.0000000000 2.85 14.0000000000 21.0000000000\\
11.0000000000 3.85 11.0000000000 44.0000000000\\
27.0000000000 4.85 27.0000000000 38.0000000000\\
11.0000000000 5.85 6.0000000000 9.0000000000\\
37.0000000000 6.85 7.0000000000 8.0000000000\\
7.0000000000 7.85 7.0000000000 8.0000000000\\
15.0000000000 8.85 15.0000000000 25.0000000000\\
10.3157894737 9.85 10.3157894737 21.2631578947\\
};
\addplot+[forget plot,only marks,mark=*,mark size=1.7pt,color=tsgcolor,
mark options={fill=tsgcolor,draw=tsgcolor,line width=0.6pt},
error bars/.cd,x dir=both,x explicit,error mark=|,
error bar style={draw=tsgcolor!65,line width=0.55pt},
error mark options={rotate=0,draw=tsgcolor,mark size=2pt,line width=0.6pt}
] table[x=mean,y=row,x error minus=minus,x error plus=plus,row sep=\\] {
mean row minus plus\\
66.0000000000 0.15 16.0000000000 14.0000000000\\
71.0000000000 1.15 11.0000000000 14.0000000000\\
43.0000000000 2.15 23.0000000000 32.0000000000\\
28.0000000000 3.15 8.0000000000 12.0000000000\\
23.0000000000 4.15 23.0000000000 27.0000000000\\
38.0000000000 5.15 28.0000000000 42.0000000000\\
21.0000000000 6.15 16.0000000000 14.0000000000\\
46.0000000000 7.15 6.0000000000 19.0000000000\\
9.0000000000 8.15 9.0000000000 16.0000000000\\
17.0000000000 9.15 12.0000000000 13.0000000000\\
12.2631578947 10.15 12.2631578947 14.0526315789\\
};
\addlegendimage{legend image code/.code={
  \draw[draw=basecolor,line width=0.6pt] (0cm,0cm) -- (0.45cm,0cm);
  \draw[draw=basecolor,line width=0.6pt] (0cm,-0.07cm) -- (0cm,0.07cm)
    (0.45cm,-0.07cm) -- (0.45cm,0.07cm);
  \draw[draw=basecolor,fill=white,line width=0.6pt]
    plot[mark=square*,mark size=1.7pt] coordinates {(0.225cm,0cm)};
}}
\addlegendentry{CFG-only}
\addlegendimage{legend image code/.code={
  \draw[draw=tsgcolor,line width=0.6pt] (0cm,0cm) -- (0.45cm,0cm);
  \draw[draw=tsgcolor,line width=0.6pt] (0cm,-0.07cm) -- (0cm,0.07cm)
    (0.45cm,-0.07cm) -- (0.45cm,0.07cm);
  \draw[draw=tsgcolor,fill=tsgcolor,line width=0.6pt]
    plot[mark=*,mark size=1.7pt] coordinates {(0.225cm,0cm)};
}}
\addlegendentry{TSG}
\end{axis}
\end{tikzpicture}
      \end{subfigure}\hfill
      \begin{minipage}[t]{0.50\textwidth}
          \centering
          \vspace{0pt}
          \begin{subfigure}[t]{\linewidth}
              \centering
              % Requires \usepackage{pgfplots} and \pgfplotsset{compat=1.18}.
% Same accuracy-matched 25 examples per model and setting.
% Measured rho values only: 0, 1, 2, 3, 4, 5. Lines connect observations.
% Pass counts: Wan2.1 = 4,8,6,5,5,4; Wan2.2 = 6,9,8,7,6,5 (out of 25).
% Input at the intended column/subfigure width; do not resize the text.
\begin{tikzpicture}
\definecolor{rhoWanOne}{HTML}{C47743}
\definecolor{rhoWanTwo}{HTML}{237E83}
\begin{axis}[
  width=\linewidth,
  height=3.5cm,
  xmin=-0.15,xmax=5.15,
  ymin=0,ymax=40,
  xtick={0,1,2,3,4,5},
  ytick={0,10,20,30,40},
  xlabel={TSG strength $\rho$ ($0$: CFG-only)},
  ylabel={Sat Acc. (\%)},
  font=\normalfont\fontsize{8}{9}\selectfont,
  tick label style={font=\normalfont\fontsize{8}{9}\selectfont},
  label style={font=\normalfont\fontsize{8}{9}\selectfont},
  axis x line*=bottom,axis y line*=left,
  axis line style={draw=gray!65,line width=0.4pt},
  tick align=outside,major tick length=2pt,
  tick style={draw=gray!65,line width=0.4pt},
  ymajorgrids=true,grid style={draw=gray!18,line width=0.3pt},
  legend style={at={(0.5,1.03)},anchor=south,
    draw=none,fill=none,legend columns=2,
    cells={anchor=west},column sep=6pt,
    font=\normalfont\fontsize{8}{9}\selectfont},
]
\addplot[color=rhoWanOne,line width=0.9pt,mark=square*,mark size=2pt,
  mark options={fill=white,draw=rhoWanOne,line width=0.8pt}]
  coordinates {(0,16) (1,32) (2,24) (3,20) (4,20) (5,16)};
\addlegendentry{Wan2.1 1.3B}
\addplot[color=rhoWanTwo,line width=0.9pt,dashed,mark=*,mark size=2pt,
  mark options={solid,fill=rhoWanTwo,draw=rhoWanTwo}]
  coordinates {(0,24) (1,36) (2,32) (3,28) (4,24) (5,20)};
\addlegendentry{Wan2.2 5B}
\end{axis}
\end{tikzpicture}
          \end{subfigure}

          \vspace{2pt}

          \begin{subfigure}[t]{\linewidth}
              \centering
              % Requires \usepackage{pgfplots} and \pgfplotsset{compat=1.18}.
% Input inside the intended column/subfigure; inherits the paper's font family.
\begin{tikzpicture}
\definecolor{cfgWanOne}{HTML}{C47743}
\definecolor{cfgWanTwo}{HTML}{237E83}
\begin{axis}[
  width=\linewidth, height=3cm,
  xmin=-0.5, xmax=4.5, ymin=0, ymax=42,
  ytick={0,10,20,30,40},
  xtick={0,1,2,3,4},
  xticklabels={CFG,CFG ($2\times$),TSG-only,TSG $\rightarrow$ CFG,CFG + TSG},
  xticklabel style={rotate=30,anchor=north east,font=\normalfont\fontsize{7}{8}\selectfont},
  ylabel={Sat Acc. (\%)},
  font=\normalfont\fontsize{8}{9}\selectfont,
  tick label style={font=\normalfont\fontsize{8}{9}\selectfont},
  label style={font=\normalfont\fontsize{8}{9}\selectfont},
  axis x line*=bottom, axis y line*=left,
  axis line style={draw=gray!65,line width=0.4pt},
  tick align=outside, major tick length=2pt,
  tick style={draw=gray!65,line width=0.4pt},
  xtick style={draw=none},
  ymajorgrids=true, grid style={draw=gray!18,line width=0.3pt},
  axis on top=false,
  legend style={at={(0.5,1.03)},anchor=south,
    draw=none,fill=none,legend columns=2,
    cells={anchor=west},column sep=5pt,
    font=\normalfont\fontsize{8}{9}\selectfont},
  legend image code/.code={
    \path[#1,draw=none] (0cm,-0.07cm) rectangle (0.24cm,0.07cm);
  },
]
\addplot[ybar,bar width=7pt,bar shift=-4.5pt,
  draw=none,fill=cfgWanOne]
  coordinates {(0,16) (1,24) (2,8) (3,12) (4,32)};
\addlegendentry{Wan2.1 1.3B}
\addplot[ybar,bar width=7pt,bar shift=4.5pt,
  draw=none,fill=cfgWanTwo]
  coordinates {(0,24) (1,20) (2,24) (3,28) (4,36)};
\addlegendentry{Wan2.2 5B}
\end{axis}
\end{tikzpicture}
          \end{subfigure}
      \end{minipage}
    \vspace{-0.1in}
      \caption{\textbf{Category-wise results and ablations.}
      Left: mean satisfaction accuracy across five models, with min--max whiskers.
      Right: guidance-strength (top) and CFG--TSG (bottom) ablations.}
      \vspace{-0.15in}
      \label{fig:TSG-ablations}
  \end{figure*}

\paragraph{Category-wise results and qualitative examples.}
Figure~\ref{fig:TSG-ablations} (left) shows category-wise satisfaction accuracies, averaged across five models, with whiskers indicating the minimum and maximum. TSG improves mean accuracy across all 11 categories by approximately 2--43 percentage points. The largest gains occur for alternative final states (23\% to 66\%), two-state transitions (35\% to 71\%), and reversible state returns (21\% to 43\%). Performance nevertheless varies substantially across models: with TSG, accuracy ranges from 20\% to 75\% for reversible state returns and from 10\% to 80\% for transitions with a persistent entity. Handover with overlap, triggered responses, and three ordered entries remain challenging, with mean accuracies of approximately 9\%, 17\%, and 12\%, respectively, and gains of only about 2 percentage points. These results show that TSG is particularly effective for state-transition requirements, while coordinating multiple entity appearances, disappearances, and dependencies remains challenging.

\begin{figure}[!htbp]
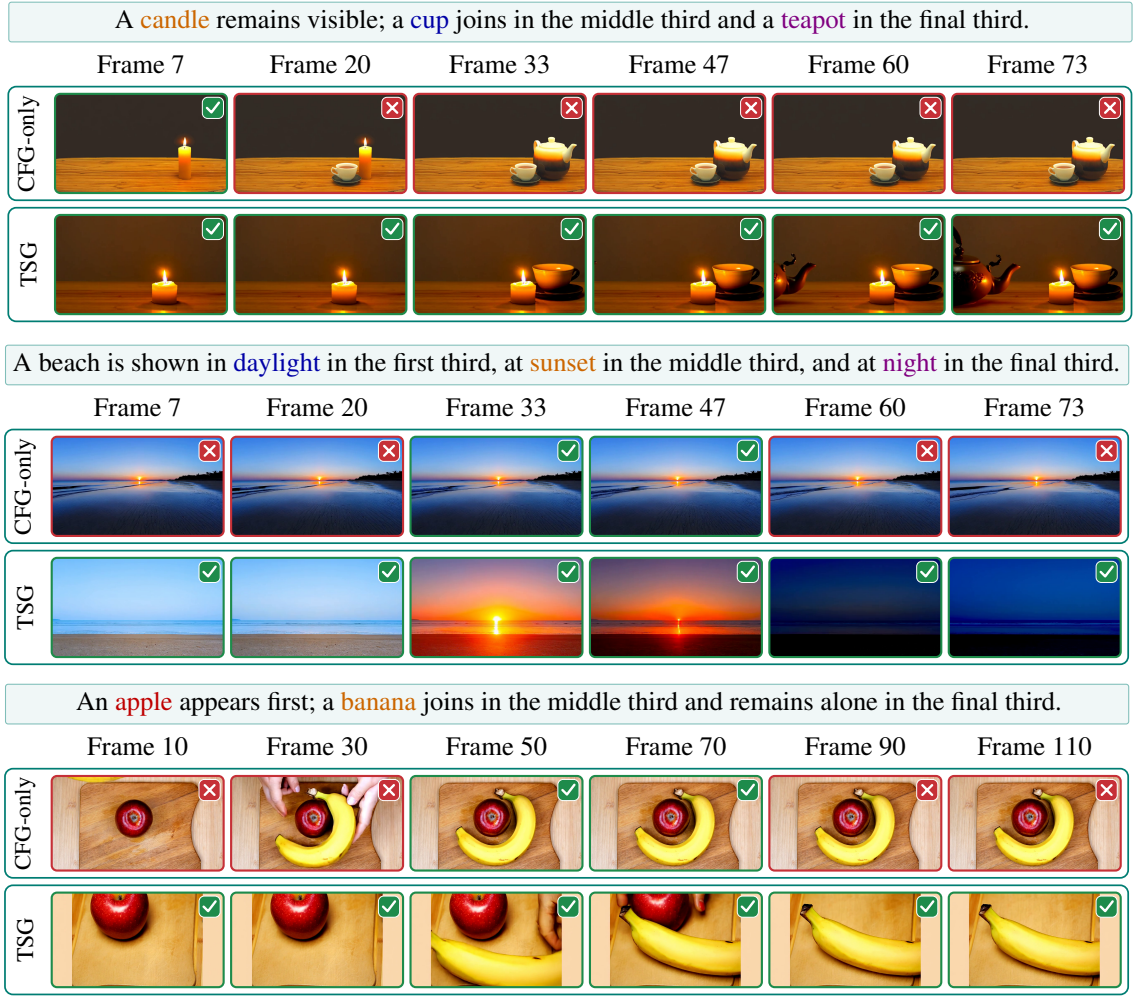

      \centering

      \begin{subfigure}{\linewidth}
          \centering
          \include{ICLR_2027/figures/examples/candle_cup_teapot/figure}
         
      \end{subfigure}

      \vspace{-0.2in}

      \begin{subfigure}{\linewidth}
          \centering
          
          \input{ICLR_2027/figures/examples/beach_day_sunset_night/figure}
      \end{subfigure}

      \vspace{0.6em}

      \begin{subfigure}{\linewidth}
          \centering
          
        \input{ICLR_2027/figures/examples/apple_to_banana/figure}
      \end{subfigure}

      \caption{Representative examples comparing CFG-only sampling (top row of each panel) with TSG (bottom row) on ordered entry, three-state transition, and overlapping handover respectively. The first two examples are generated using Wan2.1 T2V-1.3B, and the third example is generated using Wan2.2 TI2V-5B in text-to-video mode.}
      \vspace{-0.1in}
      \label{fig:representative-examples}
  \end{figure}

Figures~\ref{fig:motivating-examples}, \ref{fig:TSG}, and \ref{fig:representative-examples} illustrate how TSG improves adherence over CFG-only sampling across temporal requirements from various categories in Figure~\ref{fig:benchmark-categories}. 

\paragraph{Varying the TSG Guidance strength.}
We vary the TSG strength \(\rho \in \{0,1,2,3,4,5\}\) on a fixed
25-prompt subset per model (top-right in Figure~\ref{fig:TSG-ablations}), with \(\rho=0\) corresponding to CFG-only sampling.
Both models attain their highest satisfaction accuracy at \(\rho=1\): \(32\%\) for Wan2.1 1.3B and \(36\%\) for Wan2.2 5B, improving over CFG-only by 16 and 12 percentage points, respectively.
The accuracy is non-increasing beyond \(\rho=1\), and at \(\rho=5\) it matches or falls below the respective CFG-only baseline.
These results support \(\rho=1\) as the default among the tested strengths on this subset and show that increasing specification guidance beyond this level does not improve adherence. Appendix~\ref{appendix:ablations} provides the sampling configurations and ablation protocol.

\paragraph{TSG-CFG interaction.}
To evaluate whether the gains from TSG are an artifact of a higher guidance strength alone, we compare it against CFG with twice the guidance strength (bottom-right in Figure~\ref{fig:TSG-ablations}). We further evaluate the need for the additional prompt-conditional pass by comparing our hybrid CFG + TSG setup against TSG only, and TSG only for the first 15 denoising steps followed by CFG.
Combined CFG + TSG reports the highest specification satisfaction accuracies of \(32\%\) and \(36\%\) on Wan2.1 1.3B and Wan2.2 5B, respectively.
Doubling the CFG strength yields \(24\%\) and \(20\%\), respectively, falling short of combined guidance on both models. The effects of replacing CFG with TSG are mixed across models.
These results support our design choice of combining CFG with TSG. 

\section{Related work}
\label{sec:related-work}

\paragraph{Test-time control for video generation.}
Methods for controllable video generation differ primarily in terms of the \textit{control surface} and the \textit{time of control} (during or post sampling). Control surfaces include spatiotemporal scene layouts~\citep{lvd,ttom}, and motion transfer using reference videos~\citep{motionclone}, which are complementary to TSG's focus on temporal constraints over frame-level visual concepts. Closest to our work, TempoControl~\citep{schiber2026tempocontrol} aligns concept attention traces with prescribed binary temporal masks by repeatedly optimizing video latents during sampling. TSG offers control over more expressive temporal constraints and outperforms TempoControl on strict-timed traces with lower computational overhead (Table~\ref{tab:tempocontrol-results}). Post-sampling temporal control methods include NeuS-E~\citep{choi2025fixinpost} that uses Vision Language Models (VLM) to detect visual concepts and feedback from a Temporal Logic Model Checker to edit faulty video frames. TSG and NeuS-E offer tradeoffs as TSG does not require access to a VLM or Image Editing model, while NeuS-E can be applied to black-box models. 
MinT~\citep{mint} learns temporal control by fine-tuning a video diffusion transformer on temporally grounded data. TSG operates at test-time with frozen parameters and no additional training data.

\paragraph{Attention steering for text, image, and video generation.}
Attention provides both a signal of concept grounding and a mechanism for controlling generation of text~\citep{pasta,instaboost}, and images~\citep{prompttoprompt,attendandexcite}. 
VideoTetris~\citep{videotetris} composes attention maps across space and time to control spatial positions of concepts. 
TSG supports richer temporal compositions beyond transitions but does not directly support control of spatial positions.

\paragraph{Benchmarks for video generation.}
VBench~\citep{vbench} assesses visual quality, temporal consistency, and semantic alignment, while T2V-CompBench~\citep{t2vcompbench} evaluates compositional capabilities such as attribute binding, spatial relationships, actions, and motion.
These broader evaluations remain complementary to specification satisfaction.
TC-Bench~\citep{tcbench} and NeuS-V~\citep{neusv}  evaluate untimed state transitions and temporal relationships, while TempoControl~\citep{schiber2026tempocontrol} tests exact timing of object appearance. These temporal constraints complement the \textit{interval-timed} constraints in our benchmark. Our benchmark additionally evaluates alternate outcomes through choice, not supported by any prior benchmark.

\section{Conclusion and Future Work}
\label{sec:conclusion}

We present Temporal Specification Guidance (TSG), a training-free framework that turns temporal specifications into executable objectives for guiding video generation during sampling. We instantiate TSG with attention-derived concept traces, differentiable Metric Temporal Logic scores, and attention interventions that avoid backpropagation through the video transformer. This instantiation improves specification adherence across five models on a 220-prompt benchmark spanning 11 categories and achieves comparable temporal accuracy to TempoControl at lower computational cost.

Future work should explore traces of concept evidence beyond video-text attention maps, along with other scoring functions. TSG is limited to the latent temporal resolution of models and extending it to finer-grained temporal constraints is an interesting direction, along with identifying richer specification languages to express temporal requirements. 

\subsection*{AI use statement}

In this work, we used generative AI tools to assist benchmark construction, including proposing entities and states and drafting prompts and specification instances (Appendix~\ref{appendix:benchmark}), and to assist manuscript editing and LaTeX preparation.
We have reviewed all AI-assisted work and take responsibility for the final
content of this work, including text, claims, and artifacts produced with the
aid of generative AI.

\subsection*{Reproducibility statement}

We present details on benchmark construction, along with specification templates and example prompts, in
Appendix~\ref{appendix:benchmark}.
We further document settings used with different video generation models, along with detailed evaluation protocol in
Appendix~\ref{appendix:experimental-setup}. Finally, we present pseudocode for the TSG algorithm in Appendix~\ref{appendix:tsg-algorithm} and the quantitative semantics in 
Appendix~\ref{appendix:logstop}.

\nocite{*}
\bibliographystyle{ICLR_2027/iclr2027_conference}
\bibliography{ICLR_2027/refs}

\clearpage
\appendix

\section{Boolean MTL semantics and examples}
\label{appendix:mtl_semantics}
Given the frame-level labels \(y(v_i, c)\) for visual concepts, the label \(y(v, \varphi, t)\) for a video \(v\) satisfying specification \(\varphi\) starting from frame \(1 \leq t \leq T\) can be computed deterministically using the standard Boolean semantics for \(MTL_f\) as follows:

\begin{itemize}
    \item \(y(v,\top,t)=1\) and \(y(v,\bot,t)=0\).

    \item \(y(v,c,t)=y(v_t,c)\).

    \item \(y(v,\neg c,t)=1\) iff \(y(v,c,t)=0\).

    \item \(y(v,\varphi_1\land\varphi_2,t)=1\) iff
    \(y(v,\varphi_1,t)=1\) and \(y(v,\varphi_2,t)=1\).

    \item \(y(v,\varphi_1\lor\varphi_2,t)=1\) iff
    \(y(v,\varphi_1,t)=1\) or \(y(v,\varphi_2,t)=1\).

    \item \(y(v,\mathbf X_I\varphi,t)=1\) iff \(t<T\), \(a_I\leq 1\leq b_I\), and \(y(v,\varphi,t+1)=1\)

    \item \(y(v,\mathbf G_I\varphi,t)=1\) iff
    \(y(v,\varphi,t')=1\) for every integer \(t'\) satisfying
    \(t+a_I\leq t'\leq\min\{t+b_I,T\}\).

    \item \(y(v,\varphi_1\mathbf U_I\varphi_2,t)=1\) iff
    there exists an integer \(t'\) satisfying
    \(t+a_I\leq t'\leq\min\{t+b_I,T\}\) such that
    \(y(v,\varphi_2,t')=1\) and
    \(y(v,\varphi_1,t'')=1\) for every integer \(t''\)
    satisfying \(t\leq t''<t'\).
\end{itemize}

where \([a_I, b_I]\) is the integer-offset interval corresponding to \(I = [t_1, t_2]\), with \(a_I=\left\lceil t_1(T-1)\right\rceil\) and \(b_I=\left\lfloor t_2(T-1)\right\rfloor\).

\paragraph{Example: departure with different timing constraints.}
For instance, we can represent "a dog leaves the scene" in a 5-s long video, with decreasing flexibility in timing as:

\[
\underbrace{
  \mathsf{dog}\,\mathbf{U}\,(\mathbf{G}\neg\mathsf{dog})
}_{\textcolor{black!60}{\text{Leaves at an unspecified time}}}
\qquad
\underbrace{
  \mathsf{dog}\,\mathbf{U}_{[0.5,1]}\,
  (\mathbf{G}\neg\mathsf{dog})
}_{\textcolor{black!60}{\text{Leaves at some point during the second half}}}
\qquad
\underbrace{
  \mathsf{dog}\,\mathbf{U}_{[0.6,0.6]}\,
  (\mathbf{G}\neg\mathsf{dog})
}_{\textcolor{black!60}{\text{Leaves at the 3s mark}}}.
\]

\section{Quantitative semantics for MTL}
\label{appendix:logstop}
\label{appendix:scoring-function}

We define a differentiable sum--product score for the fragment of \(MTL_f\) in Section~\ref{sec:problem-setup}. Inspired by the log-domain computation of LogSTOP~\citep{logstop}, this construction aggregates support for the witnesses permitted by a specification. The score need not lie in \([0,1]\) and is not interpreted as a probability. In particular, disjunction and existential temporal operators sum support for alternatives, rather than combining them through complements of composite scores.

\subsection{Score definition and evaluation}
\label{appendix:scoring-definition}

Fix a sampling step, attention layer, and head, and write \(p_{t,c}=\hat y(z_t,c)\in[0,1]\) for the calibrated concept scores at latent frames \(t=1,\ldots,N\), where \(N=T_z\). Let \(Q_\psi[t]\geq0\) denote the score of subformula \(\psi\) starting at frame \(t\), and let \(S_\psi[t]=\log Q_\psi[t]\), with \(\log0=-\infty\). The scalar objective for steering is \(\hat y(z,\varphi)=S_\varphi[1]\).

For a normalized interval \(I=[\ell_I,u_I]\), define
\begin{equation}
 \begin{aligned}
 a_I&=\left\lceil\ell_I(N-1)\right\rceil,\qquad
 b_I=\left\lfloor u_I(N-1)\right\rfloor,\\
 J_I(t)&=\{j\in\mathbb Z:t+a_I\leq j\leq\min(t+b_I,N)\}.
 \end{aligned}
 \label{eq:score-window}
\end{equation}
An omitted interval denotes \([0,1]\). Every interval is relative to the evaluation frame of its operator; the conversion to offsets always uses the full trace length \(N\), including for nested operators. The set \(J_I(t)\) can be empty.

The literal scores are \(Q_c[t]=p_{t,c}\) and \(Q_{\neg c}[t]=1-p_{t,c}\), with \(Q_\top[t]=1\) and \(Q_\bot[t]=0\). Conjunction and Always multiply their operand scores, whereas disjunction and Eventually sum them. Next selects the score at \(t+1\) if this frame exists and \(1\in[a_I,b_I]\), and is zero otherwise. Until is defined by
\begin{equation}
 Q_{\alpha\mathbf U_I\beta}[t]
 =\sum_{j\in J_I(t)}
 Q_\beta[j]\prod_{r=t}^{j-1}Q_\alpha[r].
 \label{eq:logstop-until}
\end{equation}
Each summand represents a witness at \(j\): the left operand is supported at every preceding frame, and the right operand is supported at \(j\). There is no requirement that the right operand fail before \(j\). In particular, \(Q_{\mathbf F_I\beta}[t]=\sum_{j\in J_I(t)}Q_\beta[j]=Q_{\top\mathbf U_I\beta}[t]\).

For integer offsets \(0\leq a\leq b\), write \(U_{a,b}(t)=Q_{\alpha\mathbf U_{[a,b]}\beta}[t]\). Equation~\ref{eq:logstop-until} also gives the recurrence
\begin{equation}
 U_{a,b}(t)=
 \begin{cases}
 Q_\alpha[t]U_{a-1,b-1}(t+1), & a>0,\\
 Q_\beta[t]+Q_\alpha[t]U_{0,b-1}(t+1), & a=0,
 \end{cases}
 \label{eq:score-until-recurrence}
\end{equation}
where a remainder with no valid witness has score zero. Thus the recurrence inside the permitted window is \(B+AR\), where \(A\), \(B\), and \(R\) are the left-operand, right-operand, and remaining Until scores, respectively.

\subsection{Linear-time evaluation}
\label{appendix:scoring-complexity}

Evaluating Equation~\ref{eq:logstop-until} independently at every starting frame would repeat work across overlapping windows. Instead, we compute all windows using associative summaries. For fixed child-score arrays \(A_t=Q_\alpha[t]\) and \(B_t=Q_\beta[t]\), summarize a consecutive block \([i,j]\) by
\begin{equation}
 P_{i:j}=\prod_{r=i}^{j}A_r,\qquad
 R_{i:j}=\sum_{k=i}^{j}B_k\prod_{r=i}^{k-1}A_r.
 \label{eq:score-block-summary}
\end{equation}
The first component is the left-operand product, and the second is the weighted witness sum starting at the block's first frame. An empty block has summary \((1,0)\).

\begin{mylemma}[Composition of consecutive blocks]
\label{lem:score-block-composition}
Summaries of consecutive blocks combine in temporal order under the associative operation
\begin{equation}
 (P_1,R_1)\circ(P_2,R_2)
 =(P_1P_2,\ R_1+P_1R_2),
 \label{eq:score-block-combine}
\end{equation}
with identity \((1,0)\).
\end{mylemma}

\begin{proof}
The left-operand product over the concatenation is \(P_1P_2\). A witness either lies in the first block, contributing \(R_1\), or lies in the second and requires the left operand throughout the first, contributing \(P_1R_2\). Both parenthesizations of three summaries give \((P_1P_2P_3,R_1+P_1R_2+P_1P_2R_3)\), establishing associativity. The identity follows by substitution. The argument allows zero scores and does not require division.
\end{proof}

For a nonempty witness window, the Until score factors as
\begin{equation}
 Q_{\alpha\mathbf U_I\beta}[t]
 =\left(\prod_{r=t}^{t+a_I-1}A_r\right)
 R_{t+a_I:\min(t+b_I,N)}.
 \label{eq:score-until-window-factorization}
\end{equation}
The prefix maintains the left operand before the window; the block summary handles all witnesses inside it. Both factors can be obtained for all starting frames by sliding windows. An empty witness window still has score zero.

We compute block summaries in log space. For \((u,v)=(\log P,\log R)\), Equation~\ref{eq:score-block-combine} becomes
\begin{equation}
 (u,v)\star(u',v')
 =\bigl(u+u',\operatorname{logaddexp}(v,u+v')\bigr),
 \label{eq:score-log-block-combine}
\end{equation}
with identity \((0,-\infty)\). A single frame has summary \((S_\alpha[t],S_\beta[t])\). This avoids direct products of scores and differences of potentially infinite prefix sums.

\begin{algorithm}[t]
\caption{Linear-time interval-aware sum--product scoring in log space}
\label{alg:logstop-mtl}
\begin{algorithmic}[1]
\Require Concept scores \(p_{t,c}=\hat y(z_t,c)\in[0,1]\), \(t=1,\ldots,N=T_z\); specification \(\varphi\)
\Ensure Log score \(S_\varphi[1]=\hat y(z,\varphi)\) for a fixed sampling step, layer, and head
\State Convert each normalized interval to \((a_I,b_I)\) as in Equation~\ref{eq:score-window}.
\Statex \(\Call{WindowFold}{x,a,b,\circ,e}\) returns all window aggregates (Algorithm~\ref{alg:score-window-fold}).
\Statex \(\star\) is the log-domain block operation in Equation~\ref{eq:score-log-block-combine}.
\For{each distinct subformula \(\psi\) in postorder (children before parents)}
    \If{\(\psi=\mathbf G_I\alpha\)}
        \State \(S_\psi\gets\Call{WindowFold}{S_\alpha,a_I,b_I,+,0}\)
    \ElsIf{\(\psi=\mathbf F_I\alpha\)}
        \State \(S_\psi\gets\Call{WindowFold}{S_\alpha,a_I,b_I,\operatorname{logaddexp},-\infty}\)
    \ElsIf{\(\psi=\alpha\,\mathbf U_I\,\beta\)}
        \State \(H\gets\Call{WindowFold}{S_\alpha,0,a_I-1,+,0}\) \Comment{Left operand before the window}
        \State \(x_j\gets(S_\alpha[j],S_\beta[j])\) for \(j=1,\ldots,N\)
        \State \(W\gets\Call{WindowFold}{x,a_I,b_I,\star,(0,-\infty)}\)
        \State \(S_\psi[t]\gets H[t]+W[t]_2\) for \(t=1,\ldots,N\)
    \Else
        \For{\(t=1,\ldots,N\)}
            \State \(\displaystyle S_\psi[t]\gets
            \begin{cases}
                0 & \psi=\top,\\
                -\infty & \psi=\bot,\\
                \log p_{t,c} & \psi=c,\\
                \log(1-p_{t,c}) & \psi=\neg c,\\
                S_\alpha[t]+S_\beta[t] & \psi=\alpha\land\beta,\\
                \operatorname{logaddexp}(S_\alpha[t],S_\beta[t]) & \psi=\alpha\lor\beta,\\
                S_\alpha[t+1] & \psi=\mathbf X_I\alpha,\ t<N,\ a_I\leq1\leq b_I,\\
                -\infty & \psi=\mathbf X_I\alpha\text{ otherwise}.
            \end{cases}\)
        \EndFor
    \EndIf
    \State Cache \(S_\psi[1:N]\).
\EndFor
\State \Return \(S_\varphi[1]\)
\Statex \(W[t]_2\) denotes the second component; each \(S_\psi\), \(H\), and \(W\) is a length-\(N\) array.
\Statex \(\log0=-\infty\); empty windows return the supplied identity \(e\).
\Statex \(\operatorname{logaddexp}(x,y)=\log(e^x+e^y)\), evaluated stably.
\end{algorithmic}
\end{algorithm}

\paragraph{Sliding-window aggregation.}
Algorithm~\ref{alg:score-window-fold} computes all window aggregates for an associative operation with a constant-size summary. Its queue is implemented with a front stack and a back stack. Every stack entry stores its value and a cached aggregate. The front aggregate \(G_F\) runs from the top to the bottom of the front stack; the back aggregate \(G_B\) runs from the bottom to the top of the back stack. An empty stack has aggregate \(e\). The operations are:
\begin{itemize}
 \item \(\operatorname{PushBack}(x)\): push \((x,G_B\circ x)\) onto the back stack.
 \item \(\operatorname{PopFront}()\): if the front stack is empty, transfer all values from the back stack, popping each value \(x\) and pushing \((x,x\circ G_F)\) onto the front stack. Then pop the front stack.
 \item \(\operatorname{Aggregate}()\): return \(G_F\circ G_B\).
\end{itemize}
The order of combination is essential because the Until operation is not commutative. Each value enters the back stack once, transfers to the front stack at most once, and leaves the queue at most once. Push and aggregate operations have constant cost; removal has amortized constant cost. No inverse operation is needed, so zero child scores and empty windows are handled directly.

\begin{algorithm}[!b]
\caption{All sliding-window aggregates in linear time}
\label{alg:score-window-fold}
\begin{algorithmic}[1]
\Require Sequence \(x_1,\ldots,x_N\); integer offsets \(a\geq0,b\); associative operation \(\circ\) with identity \(e\)
\Ensure \(W[t]=x_{t+a}\circ\cdots\circ x_{\min(t+b,N)}\), or \(e\) for an empty window
\Function{WindowFold}{$x,a,b,\circ,e$}
    \If{\(a>b\)}
        \State \Return \([e,\ldots,e]\) \Comment{Includes the empty prefix when \(a=0,b=-1\)}
    \EndIf
    \State Initialize an empty aggregate queue \(D\) with operation \(\circ\) and identity \(e\).
    \State \(l\gets1\); \(r\gets0\)
    \For{\(t=1,\ldots,N\)}
        \State \(L\gets\min(t+a,N+1)\); \(R\gets\min(t+b,N)\)
        \While{\(r<R\)}
            \State \(r\gets r+1\); \(D.\operatorname{PushBack}(x_r)\)
        \EndWhile
        \While{\(l<L\)}
            \State \(D.\operatorname{PopFront}()\); \(l\gets l+1\)
        \EndWhile
        \State \(W[t]\gets D.\operatorname{Aggregate}()\)
    \EndFor
    \State \Return \(W\)
\EndFunction
\Statex The queue uses two stacks with cached aggregates, as defined in Appendix~\ref{appendix:scoring-complexity}.
\Statex Each element is inserted once and removed at most once; aggregation preserves sequence order.
\end{algorithmic}
\end{algorithm}

\begin{myproposition}[Correctness and complexity of evaluation]
\label{prop:score-linear-time}
Algorithm~\ref{alg:logstop-mtl} computes the stated sum--product log score using \(O(|\varphi|N)\) arithmetic operations and \(O(|\varphi|N)\) storage, including for nested temporal operators. These bounds count operations on scalar scores as constant-time operations.
\end{myproposition}

\begin{proof}
For fixed offsets \(a\leq b\), both endpoints of the clipped window advance monotonically with \(t\). Algorithm~\ref{alg:score-window-fold} therefore inserts each sequence element once and removes it at most once. The cached stack aggregates preserve the sequence order by associativity, so each returned value is the requested window aggregate. Including transfers between stacks, the total work for all \(N\) windows is \(O(N)\). If \(a>b\), the algorithm directly returns the identity at every frame in \(O(N)\) time.

Always uses addition of log scores with identity zero; Eventually uses log-add-exp with identity \(-\infty\). Until uses one window computation for the prefix log product and one for the log-domain block summaries. Lemma~\ref{lem:score-block-composition} and Equation~\ref{eq:score-until-window-factorization} show that their combination equals Equation~\ref{eq:logstop-until}, including empty windows. Constants, literals, Boolean operators, and Next take \(O(N)\) work by pointwise evaluation. Processing subformulas in postorder proves correctness by induction and evaluates each distinct subformula once, giving \(O(|\varphi|N)\) total work. Cached score arrays require \(O(|\varphi|N)\) storage, and the working arrays and aggregate queues require \(O(N)\) additional space per subformula. The argument depends only on the number of subformulas, not on their nesting depth.
\end{proof}

The computation also has \(O(|\varphi|N)\) scalar intermediate operations. At differentiable inputs, reverse-mode differentiation can therefore use the same asymptotic time and storage bounds, with structurally zero branches handled as constants. This is an arithmetic complexity statement; it does not establish a wall-clock speedup over a vectorized quadratic implementation on a GPU.

\paragraph{Numerical evaluation.}
The mathematical definition admits exact zeros and ones. For numerical gradients, literal inputs can be clipped to \([\epsilon_{\mathrm{LS}},1-\epsilon_{\mathrm{LS}}]\), separately from the calibration in Equation~\ref{eq:TSG-trace}. Composite scores are not clipped to one, and negation is applied only to atoms. Products are computed as sums of logarithms, and sums as log-sum-exp reductions. Empty products are one and empty sums are zero: Always on an empty window has log score zero, while Eventually and Until have log score \(-\infty\). On the domain where \(Q_\varphi[1]>0\), the log score is differentiable for literal inputs strictly between zero and one. Structurally zero branches, including log-add-exp with two constant \(-\infty\) inputs, should be handled explicitly when differentiating. If the entire score is identically zero, for example because a required witness window is empty, its logarithm provides no finite gradient; the specification must be checked for feasibility on the chosen grid, and that objective should not produce a steering update.

\subsection{Monotonicity and its scope}
\label{appendix:scoring-monotonicity}

To distinguish support for a literal from the underlying concept score, introduce independent nonnegative inputs \(x^+_{t,c}\) and \(x^-_{t,c}\) for \(c\) and \(\neg c\), respectively. Write \(\mathcal Q_\varphi(x)\) for the score obtained from the same rules with these literal inputs. The score on a concept trace is recovered by substituting \(x^+_{t,c}=p_{t,c}\) and \(x^-_{t,c}=1-p_{t,c}\).

\begin{samepage}
\begin{myproposition}[Monotonicity in literal support]
\label{prop:score-monotonicity}
Fix a finite temporal grid and the intervals of a specification \(\varphi\) in the stated fragment. The function \(\mathcal Q_\varphi\) is a polynomial with nonnegative coefficients in the independent literal-support inputs. Consequently, for any componentwise ordered inputs \(0\leq x\leq x'\),
\[
 \mathcal Q_\varphi(x)\leq\mathcal Q_\varphi(x').
\]
The same inequality holds for their logarithms wherever both scores are positive. The claim allows arbitrary nesting of the supported operators and is non-strict.
\end{myproposition}
\end{samepage}

\begin{proof}
We proceed by structural induction on the formula, simultaneously for every starting frame. Constants are nonnegative constant polynomials, and literal scores are individual input variables. Conjunction and Always take finite products of child scores; disjunction and Eventually take finite sums. These operations preserve polynomials with nonnegative coefficients, including their empty-product and empty-sum conventions. Next selects a child score or the constant zero. Finally, each Until score in Equation~\ref{eq:logstop-until} is a finite sum of products of child scores, and therefore has the same property. This completes the induction. Every monomial with a nonnegative coefficient is nondecreasing in nonnegative inputs, so their sum is nondecreasing as well. The logarithm is increasing on the positive real numbers.
\end{proof}

\begin{mycorollary}[Monotonicity in a concept--time coordinate]
\label{cor:score-polarity}
Expand the score at a fixed starting frame into literal-support inputs. If a concept--time pair \((t,c)\) occurs only positively, the specification score is nondecreasing in \(p_{t,c}\); if it occurs only negatively, the specification score is nonincreasing in \(p_{t,c}\), with all other concept scores fixed. These directions also hold for its logarithm wherever the specification score is positive.
\end{mycorollary}

\begin{proof}
In the first case, increasing \(p_{t,c}\) increases the only relevant literal input \(x^+_{t,c}\). In the second, it decreases the only relevant input \(x^-_{t,c}=1-p_{t,c}\). Apply Proposition~\ref{prop:score-monotonicity}.
\end{proof}

Consistency of polarity is required only for the concept--time coordinate in this sufficient condition, not for an entire concept throughout the formula. For example, presence in an early interval and absence in a disjoint later interval have consistent directions at each frame. Both polarities are allowed at the same frame, but then
\begin{equation}
 \frac{\partial Q_\varphi[1]}{\partial p_{t,c}}
 =\frac{\partial\mathcal Q_\varphi}{\partial x^+_{t,c}}
  -\frac{\partial\mathcal Q_\varphi}{\partial x^-_{t,c}},
 \label{eq:score-mixed-polarity}
\end{equation}
evaluated at the corresponding literal inputs. Each term on the right is nonnegative; their difference need not have a fixed sign. A formula with mixed polarity can nevertheless be monotone after simplification, as the untimed Until example below demonstrates. No probabilistic independence assumption is needed for these algebraic claims.

\paragraph{Agreement with Boolean satisfaction.}
For exact Boolean concept inputs, \(Q_\varphi[t]>0\) if and only if the trace satisfies \(\varphi\) at \(t\) under Section~\ref{sec:problem-setup}. This follows by structural induction: a product of nonnegative scores is positive exactly when all its factors are positive, and a sum is positive exactly when at least one summand is positive. Applied to Equation~\ref{eq:logstop-until}, these conditions are precisely an admissible right-operand witness and satisfaction of the left operand at all preceding frames. The constants, literals, Next, and empty-window cases follow their Boolean definitions. This statement concerns the unclipped score; it does not prescribe a satisfaction threshold for soft inputs.

Different satisfying traces can have different scores, and logically equivalent formulas need not have identical soft scores. Summing alternatives rewards multiple supported witnesses, and repeating a disjunct changes its weight. We therefore retain the input formula's structure rather than treating Boolean rewrites as score-preserving transformations.

\subsection{Examples and implications for gradients}
\label{appendix:scoring-gradients}

For the following examples, let \(p_t=\hat y(z_t,\mathsf{dog})\in(0,1)\), and let \(u_t=\operatorname{logit}(p_t)\) be its calibrated trace logit. This \(u_t\) is distinct from a transformer attention logit. We write \(Q=Q_\varphi[1]\) and \(S=\log Q\), assuming \(Q>0\).

\paragraph{Presence and absence in prescribed intervals.}
For \(\varphi=\mathbf G_{I_+}\mathsf{dog}\land\mathbf G_{I_-}\neg\mathsf{dog}\), let \(E=J_{I_+}(1)\) and \(A=J_{I_-}(1)\) be disjoint. Then
\begin{equation}
 S=\sum_{t\in E}\log p_t+\sum_{t\in A}\log(1-p_t),\qquad
 \frac{\partial S}{\partial u_t}
 =\mathbf 1_{\{t\in E\}}(1-p_t)-\mathbf 1_{\{t\in A\}}p_t.
 \label{eq:score-fixed-gradient}
\end{equation}
Gradient ascent thus increases presence evidence in \(E\), decreases it in \(A\), and leaves unconstrained frames unaffected. A strictly timed witness is likewise obtained by restricting an existential operator to a singleton window, provided the requested time lies on the grid.

\paragraph{Untimed departure without persistence.}
For \(\varphi=\mathsf{dog}\,\mathbf U\,\neg\mathsf{dog}\), the witness sum telescopes:
\begin{equation}
 Q=\sum_{j=1}^{N}(1-p_j)\prod_{r<j}p_r
   =1-\prod_{t=1}^{N}p_t,\qquad
 \frac{\partial Q}{\partial p_t}=-\prod_{r\ne t}p_r\leq0.
 \label{eq:score-untimed-departure}
\end{equation}
Increasing absence evidence at any frame cannot decrease this score, despite its mixed polarity. Under Boolean semantics, the formula is equivalent to \(\mathbf F\neg\mathsf{dog}\): choose the first absent frame as the Until witness, and all earlier frames necessarily contain the dog. An immediate witness is permitted, and no constraint is imposed after the witness. Nevertheless, our score for \(\mathbf F\neg\mathsf{dog}\) is \(\sum_t(1-p_t)\), illustrating that Boolean equivalence does not imply equality of soft scores.

\paragraph{Departure followed by persistent absence.}
For \(\varphi=\mathsf{dog}\,\mathbf U_I\,(\mathbf G\neg\mathsf{dog})\), define
\begin{equation}
 w_j=\underbrace{\prod_{r<j}p_r}_{\text{presence before }j}\quad
     \underbrace{\prod_{r\geq j}(1-p_r)}_{\text{absence from }j},\qquad
 Q=\sum_{j\in J_I(1)}w_j,\qquad \pi_j=\frac{w_j}{Q}.
 \label{eq:score-persistent-departure}
\end{equation}
Each term corresponds to one complete departure pattern. Differentiating \(\log Q\) gives a weighted combination of the gradients of these patterns:
\begin{equation}
 \frac{\partial S}{\partial u_t}
 =\sum_{j\in J_I(1)}\pi_j\frac{\partial\log w_j}{\partial u_t}
 =\underbrace{\sum_{\substack{j\in J_I(1)\\j>t}}\pi_j}_{\text{support for being present at }t}-p_t.
 \label{eq:score-departure-gradient}
\end{equation}
Before the earliest admissible departure this derivative is \(1-p_t\); at or after the latest admissible departure it is \(-p_t\). Within the window its sign depends on the relative support for departure times on either side of \(t\). Thus the objective encourages a coherent departure, rather than absence independently at every permitted frame.

An unconditional claim that more absence always helps would contradict the required persistence. For untimed Until, the Boolean trace \([1,1,1,0]\) has score one, whereas changing its second frame to obtain \([1,0,1,0]\) gives score zero: the dog now returns after an earlier disappearance. Also, the formula as written permits absence throughout. To require initial presence, one must explicitly conjoin \(\mathsf{dog}\).

\paragraph{Nested departure and return.}
For \(\varphi=\mathsf{dog}\,\mathbf U_{I_1}(\neg\mathsf{dog}\,\mathbf U_{I_2}\mathsf{dog})\), recursive evaluation yields
\begin{equation}
 Q=\sum_{j\in J_{I_1}(1)}\sum_{k\in J_{I_2}(j)}
 \underbrace{\prod_{r<j}p_r}_{\text{before departure}}\quad
 \underbrace{\prod_{r=j}^{k-1}(1-p_r)}_{\text{absent until return}}\quad
 \underbrace{p_k}_{\text{return}}.
 \label{eq:score-nested-return}
\end{equation}
Each admissible pair \((j,k)\) contributes gradients that favor presence before \(j\), absence from \(j\) through \(k-1\), and presence at \(k\); it imposes no condition after \(k\). The log-score gradient is their weighted combination, with weights proportional to the corresponding summands. Replacing the final \(\mathsf{dog}\) by \(\mathbf G\mathsf{dog}\) replaces \(p_k\) by \(\prod_{r=k}^{N}p_r\) and requires persistence after return.

The second interval is measured from \(j\), not from the start of the video. A lower bound \(a_{I_2}\geq1\) requires at least one absent frame between departure and return. In particular, for \(N\geq2\) and \(I_1=I_2=[0.5,1]\), departure must occur at the midpoint and return at the final frame. This pair exists only if the midpoint lies on the sampled grid: writing \(d=\lceil(N-1)/2\rceil\), two successive delays require \(2d\leq N-1\). Otherwise the witness set is empty. Shorter lower bounds permit multiple departure--return pairs.

\paragraph{Alternative outcomes.}
For \(\varphi=\alpha\lor\beta\) with positive operand scores,
\begin{equation}
 \nabla S
 =\frac{Q_\alpha}{Q_\alpha+Q_\beta}\nabla\log Q_\alpha
  +\frac{Q_\beta}{Q_\alpha+Q_\beta}\nabla\log Q_\beta.
 \label{eq:score-choice-gradient}
\end{equation}
The update combines the directions for both alternatives according to their current support. A disjunction alone does not impose exclusivity or select a single branch; when exclusivity is intended, it must be encoded in the specification. Similarly, summing temporal witnesses can favor support at several admissible times rather than only one.

\paragraph{Transferring gradients to attention biases.}
For any differentiable input representation and \(Q>0\), \(\nabla S=\nabla Q/Q\). Thus the nonzero gradients of \(S\) and \(Q\) have the same direction after the infinity-norm normalization used by TSG, before numerical thresholds and clipping. The score need not be bounded by one for this normalization to apply. If the calibration statistics are held fixed while differentiating Equation~\ref{eq:TSG-trace}, then \(u_{t,c}=(a_{t,c}-\mu_c)/(\tau s_c+\epsilon)\) gives
\[
 \frac{\partial S}{\partial a_{t,c}}
 =\frac{1}{\tau s_c+\epsilon}\frac{\partial S}{\partial u_{t,c}}.
\]
The positive factor preserves the gradient signs described above.

\section{The TSG algorithm}
\label{appendix:tsg-algorithm}

\begin{algorithm}[t]
\caption{Temporal Specification Guidance with online attention steering}
\label{alg:TSG}
\begin{algorithmic}[1]
\Require Prompt embeddings $e_p,e_{\mathrm{ref}}$, specification $\varphi$, transformer $T_\theta$, decoder $D$
\Require Sampling steps $N$, guided steps $\mathcal K$, selected layer--head pairs $\mathcal H$
\Require CFG scale $\lambda_{\mathrm{cfg}}$, TSG norm ratio $\rho$, bias step size $\eta$, bounds $[b_{\min},b_{\max}]$
\Require Calibration temperature $\tau$ and stabilizer $\epsilon$
\Ensure Generated video $v$
\State Map concepts in $\varphi$ to prompt-token sets $\mathcal P_c$
\State Initialize $z_N\sim\mathcal N(0,I)$; convert intervals in $\varphi$ to the latent temporal grid
\For{$k=N,N-1,\ldots,1$}
    \State $d_{k,c}\gets T_\theta(z_k,e_p,k)$ without steering
    \State $d_{k,u}\gets T_\theta(z_k,e_{\mathrm{ref}},k)$ without steering
    \State $d_{k,\mathrm{cfg}}\gets d_{k,u}+\lambda_{\mathrm{cfg}}(d_{k,c}-d_{k,u})$
    \State $d_k\gets d_{k,\mathrm{cfg}}$
    \If{$k\in\mathcal K$}
        \State Begin an additional conditional evaluation of $T_\theta(z_k,e_p,k)$ without a transformer gradient graph
        \For{selected attention layers $\ell$, in forward order}
            \State Advance this guided pass to layer $\ell$, using outputs of the preceding layers
            \ForAll{heads $h$ with $(\ell,h)\in\mathcal H$}
                \State Compute current attention $A^{\ell,h}$ before the local intervention
                \State Form raw traces $a^{\ell,h}$ using Equation~\ref{eq:TSG-trace}; detach and enable trace gradients
                \State Calibrate to $\hat y^{\ell,h}$ using Equation~\ref{eq:TSG-trace}, holding statistics fixed
                \State $S_{\ell,h}\gets\operatorname{Score}(\hat y^{\ell,h},\varphi)$
                \ForAll{concepts $c$ in $\varphi$}
                    \State $g_{\cdot,c}\gets\nabla_{a_{\cdot,c}}S_{\ell,h}$
                    \State $b_{\cdot,c}\gets\eta g_{\cdot,c}/\|g_{\cdot,c}\|_\infty$; use zero for negligible gradients
                \EndFor
                \State Broadcast $b$ to query--text pairs; center across text keys and clip to obtain $B^{\ell,h}$
                \State Detach $B^{\ell,h}$ and apply it at this head using Equation~\ref{eq:TSG-attention-bias}
            \EndFor
            \State Complete layer $\ell$ with the biased attention; release local scoring buffers
        \EndFor
        \State $d_{k,\varphi}\gets$ output of the completed guided forward pass
        \State $r_k\gets d_{k,\varphi}-d_{k,c}$; compute $\lambda_{k,\mathrm{tsg}}$ using Equation~\ref{eq:tsg-scale}
        \State $d_k\gets d_{k,\mathrm{cfg}}+\lambda_{k,\mathrm{tsg}}r_k$
    \EndIf
    \State $z_{k-1}\gets\operatorname{Step}_k(z_k,d_k)$
\EndFor
\State \Return $D(z_0)$
\end{algorithmic}
\end{algorithm}

Algorithm~\ref{alg:TSG} presents TSG with online attention steering. The guided steps are the first \(N_{\mathrm{tsg}}\) sampling updates, i.e., \(\mathcal K=\{N-N_{\mathrm{tsg}}+1,\ldots,N\}\). The loop over selected attention layers takes place within a single additional forward pass: normal transformer operations proceed between interventions, and later layers use the representations produced by earlier steered layers. Differentiation is enabled only for the detached traces, their calibration, and the specification score.

\section{The TSG benchmark}
\label{appendix:benchmark}
\begin{table*}[!tbh]
\centering
\small
\setlength{\tabcolsep}{4pt}
\renewcommand{\arraystretch}{1.12}
\caption{The 11 temporal constraint categories in the main benchmark, each containing 20 examples. Descriptions summarize the temporal structure; individual specifications also encode timing, absence, and persistence requirements.}
\label{tab:TSG-benchmark-overview}
\begin{tabular}{@{}p{0.34\linewidth}p{0.63\linewidth}@{}}
\toprule
Category & Temporal behavior \\
\midrule
Two-state transition & A subject or setting changes from state $A$ to $B$. \\
Three-state transition & A subject or setting progresses through $A$, $B$, then $C$. \\
Ordered entry of two entities & One entity is initially present; a second joins it and both remain. \\
Ordered entry of three entities & Two further entities enter in sequence, accumulating in the scene. \\
Entity handover with overlap & $A$ alone is followed by $A$ and $B$ together, then $B$ alone. \\
Entity exit and re-entry & An entity is present, becomes absent, then returns. \\
Reversible state return & A subject changes from $A$ to $B$ and returns to $A$. \\
Anchored entity choice & An anchor persists while one of two alternative entities appears. \\
Transition with a persistent entity or setting & An anchor remains present while another concept changes state. \\
Alternative final state & An initial state gives way to either of two permitted final states. \\
Trigger-conditioned state transition & A trigger entity appears before the target changes state. \\
\bottomrule
\end{tabular}
\label{tab:tsg-categories}
\end{table*}

\paragraph{Visual concepts as states of entities.} We first use \texttt{GPT-5.6-Sol} to propose 50 entities spanning object and setting classes, together with associated states or attributes, such as color and time of day. This collection includes object classes from VBench~\citep{vbench}, augmented with states.
One author manually reviews the collection to exclude unfamiliar entities and implausible entity--state combinations. Visual concepts represent entity presence or an entity assuming a particular state. The final inventory contains 50 entries: 36 objects and 14 scenes or settings. 
Examples of object entities with states are presented in Table~\ref{tab:tsg-inventory-1} and of setting entities with states in Table~\ref{tab:tsg-inventory-3}.

\paragraph{Benchmark categories.} We manually define 11 categories of temporal constraints covering state transitions, entity appearance and disappearance, and choice. Table~\ref{tab:tsg-categories} illustrates each category. For each category, one author collaborates with \texttt{GPT-5.6-Sol} to construct 20 prompts, reviewing them for relevance, diversity of entities and states, and realism. We specify temporal intervals using halves and thirds of the video duration, allowing meaningful time for each event in a five-second video.

\begin{table}[t]
\centering
\small
\renewcommand{\arraystretch}{1.12}
\begin{tabular}{@{}p{0.19\linewidth}p{0.77\linewidth}@{}}
\hline
Entity & Candidate states or attributes \\
\hline
airplane & configuration: landed, in flight \\
apple & condition: whole, sliced \\
balloon & inflation: visibly deflated, visibly inflated \\
bicycle & orientation: upright, lying on its side \\
bird & configuration: perched, in flight \\
boat & orientation: upright, capsized \\
book & position: closed, open \\
butterfly & configuration: perched, in flight \\
campfire & flame visibility: no visible flame, visible flames \\
candle & flame: lit, unlit \\
car & orientation: upright, overturned; headlight appearance: headlights visibly on, headlights visibly off \\
cat & pose: standing, sitting, lying down \\
cup & fill level: visibly empty, filled to the rim \\
dog & pose: standing, sitting, lying down \\
door & position: closed, open \\
egg & shell: intact, cracked \\
faucet & water visibility: no visible water stream, visible water stream \\
flag & position: fully lowered, fully raised \\
\hline
\end{tabular}
\caption{Examples of object entities with corresponding states.}
\label{tab:tsg-inventory-1}
\end{table}

\begin{table}[t]
\centering
\small
\renewcommand{\arraystretch}{1.12}
\begin{tabular}{@{}p{0.19\linewidth}p{0.77\linewidth}@{}}
\hline
Entity & Candidate states or attributes \\
\hline
beach & illumination: bright daylight, orange horizon, dark night \\
city street & illumination: bright daylight, dark night; precipitation: no visible precipitation, visible falling rain, visible falling snow \\
farm field & ground cover: green crops, bare soil, snow-covered; precipitation: no visible precipitation, visible falling rain, visible falling snow \\
forest & dominant foliage: green foliage, autumn-colored foliage, completely bare branches, snow-covered trees; precipitation: no visible precipitation, visible falling rain, visible falling snow \\
kitchen & artificial lighting: lights visibly illuminated, no lights visibly illuminated \\
lake & surface: visible open water, completely ice-covered; visibility: far shore clearly visible, far shore obscured by fog; precipitation: no visible precipitation, visible falling rain, visible falling snow \\
mountain landscape & illumination: bright daylight, orange horizon, dark night; visibility: mountains clearly visible, mountains obscured by fog \\
ocean & illumination: bright daylight, orange horizon, dark night; sea appearance: calm surface, large breaking waves \\
park & illumination: bright daylight, dark night; precipitation: no visible precipitation, visible falling rain, visible falling snow \\
road & surface: visibly dry, visibly wet, snow-covered \\
room & artificial lighting: lights visibly illuminated, no lights visibly illuminated \\
sky & illumination: bright daylight, orange horizon, dark night; cloud appearance: mostly clear sky, overcast sky, dark storm clouds \\
stadium & occupancy: no visible spectators, spectators visible; floodlight appearance: floodlights visibly illuminated, floodlights visibly unlit \\
waterfall & water appearance: visible flowing water, completely frozen; surroundings: green vegetation, snow-covered ground \\
\hline
\end{tabular}
\caption{Examples of scene and setting entities with corresponding states.}
\label{tab:tsg-inventory-3}
\end{table}

\paragraph{Specification templates for different categories.} We then instantiate manually designed specification templates to encode each prompt's temporal requirements. These interval-timed specifications allow the requested transitions to occur anywhere within the designated intervals, subject to any ordering and persistence requirements. Thus, as in the examples in Section~\ref{sec:problem-setup}, they do not prescribe a frame-by-frame schedule for each concept. 

Let $T$ denote video duration. We use half intervals $H_1=[0,T/2)$ and $H_2=[T/2,T]$, or thirds $I_1=[0,T/3)$, $I_2=[T/3,2T/3)$, and $I_3=[2T/3,T]$. Thus, adjacent phases do not share a boundary frame. For a phase with positive propositions $P^+$ and negative propositions $P^-$, its predicate is
\[
 p=\bigwedge_{a\in P^+}a\;\land\!\bigwedge_{b\in P^-}\neg b.
\]
Only the listed propositions are constrained; unrelated scene content is not prohibited. A state proposition includes its entity (e.g., \emph{sitting dog}), and persistent-entity transitions retain these compound propositions rather than factoring them into separate entity and state atoms.

For compactness, write $W_J(p)$ for an eventually--always requirement: there must be a consecutive run satisfying $p$ wholly within $J$, with the steering persistence parameter set to $\delta=0.5$ seconds. In continuous notation this is $\mathbf{F}_{\{t\in J:[t,t+\delta]\subseteq J\}}\mathbf{G}_{[0,\delta]}p$; the discrete implementation is specified below. Define
\[
 S_2(p_1,p_2)=W_{H_1}(p_1)\land W_{H_2}(p_2),\qquad
 S_3(p_1,p_2,p_3)=\bigwedge_{j=1}^{3}W_{I_j}(p_j).
\]
The steering templates are:
\begin{enumerate}
\item \emph{Two-state transition (2ST):}
$S_2(A\land\neg B,B\land\neg A)$.
\item \emph{Three-state transition (TST):}
$S_3(A\land\neg B\land\neg C,B\land\neg A\land\neg C,C\land\neg A\land\neg B)$.
\item \emph{Two ordered entries (OE2):}
$S_2(A\land\neg B,A\land B)$.
\item \emph{Three ordered entries (OE3):}
$S_3(A\land\neg B\land\neg C,A\land B\land\neg C,A\land B\land C)$.
\item \emph{Handover with overlap (HWO):}
$S_3(A\land\neg B,A\land B,B\land\neg A)$.
\item \emph{Exit and return (EXR):} $S_3(A,\neg A,A)$.
\item \emph{Reversible state return (RSR):}
$S_3(A\land\neg B,B\land\neg A,A\land\neg B)$.
\item \emph{Persistent-entity transition (PST):}
$\mathbf{G}_{[0,T]}E\land S_2(A\land\neg B,B\land\neg A)$.
\item \emph{Triggered response (TRG):}
$S_3(A\land\neg B\land\neg E,A\land E\land\neg B,B\land E\land\neg A)$.
Here $E$ is the entering trigger and $A,B$ are the original and response states. This specifies an observed temporal sequence, not a causal intervention.
\item \emph{Alternative final state (AFS):}
$\mathbf{G}_{I_1}(A\land\neg B\land\neg C)\land
[\mathbf{G}_{I_3}(B\land\neg A\land\neg C)\lor
\mathbf{G}_{I_3}(C\land\neg A\land\neg B)]$.
\item \emph{Anchored entity choice (AEC):}
$\mathbf{G}_{[0,T]}E\land\mathbf{G}_{I_1}(\neg A\land\neg B)\land
[\mathbf{G}_{I_3}(A\land\neg B)\lor\mathbf{G}_{I_3}(B\land\neg A)]$.
\end{enumerate}
In the state-transition templates, $A,B,C$ refer to states of the same subject or setting; in the entry, handover, and entity-choice templates they denote entity presence. $E$ denotes an anchor or trigger as indicated. Every category has 20 examples. Choice formulas use an ordinary disjunction between \emph{exclusive branch predicates}: either branch is acceptable, but both alternatives cannot satisfy the same required frame. Placing the disjunction outside the final-region always operators selects one fixed branch throughout that region. The middle third is unconstrained by AFS steering and, apart from anchor persistence, by AEC steering; these formulas do not prohibit transient alternative states in that gap.

\paragraph{Concrete examples of prompts.}
The following are exact released prompts, with proposition assignments that instantiate the corresponding templates above. 

\begin{enumerate}
\item \emph{Two-state transition} (\texttt{2st-181\_person\_stand\_to\_sit}).
$A=$ standing person; $B=$ sitting person.

``A continuous full-body studio shot follows one person against a simple background. The person begins standing and holds that pose for roughly the first half of the video. Around the midpoint, they sit down on the floor and remain seated for the rest of the video. The same person remains in frame throughout, with no cuts, duplicate people, or camera movement.''
\item \emph{Three-state transition} (\texttt{tst-001\_traffic\_light\_yellow\_red\_green}).
$A=$ yellow traffic light; $B=$ red traffic light; $C=$ green traffic light.

``A cinematic, photorealistic 4K close-up follows one street traffic light at dusk from a fixed tripod camera. The signal is solid yellow for roughly the first third of the video. About one-third of the way through, that same signal changes cleanly to solid red. Around two-thirds of the way through, it changes to solid green and remains green through the end. The softly blurred urban background stays unchanged, with no camera movement, scene cuts, duplicate traffic lights, or extra signal flicker.''
\item \emph{Ordered entry of two entities} (\texttt{oe2-201\_person\_then\_dog}).
$A=$ person; $B=$ dog.

``One person waits alone in a quiet park clearing for roughly the first half of a continuous medium-wide shot. Around the midpoint, a dog runs into the clearing and stops nearby. The same person and dog remain fully visible together through the end, with no cuts, duplicate subjects, or camera movement.''
\item \emph{Ordered entry of three entities} (\texttt{oe3-121\_dog\_cat\_rabbit}).
$A=$ dog; $B=$ cat; $C=$ rabbit.

``One dog waits alone on a broad living-room rug for roughly the first third of a fixed shot. A cat then walks into the room and settles within the same view while the dog stays. Around two-thirds of the way through, a rabbit hops in through the open doorway. The dog, cat, and rabbit all remain visible together until the end, with no duplicate animals or camera movement.''
\item \emph{Entity handover with overlap} (\texttt{hwo-021\_horse\_to\_rabbit}).
$A=$ horse; $B=$ rabbit.

``A horse stands alone in a grassy paddock at the beginning of a steady wide shot. About one-third of the way through, a rabbit hops into the foreground while the horse remains nearby. Around the two-thirds point, the horse walks calmly out through an open gate, leaving the rabbit in view until the end. The camera does not cut or change position.''
\item \emph{Entity exit and re-entry} (\texttt{exr-061\_dog\_exit\_return}).
$A=$ dog.

``A fixed living-room shot begins with one dog standing in clear view. About one-third of the way through, the dog walks out through the left side until it is completely absent. Near the two-thirds point, the same dog walks back in from the left and remains in the room until the end. There are no cuts or other dogs.''
\item \emph{Reversible state return} (\texttt{rsr-081\_person\_stand\_sit\_stand}).
$A=$ standing person; $B=$ sitting person.

``A continuous full-body studio shot begins with one person standing against a simple background. About one-third of the way through, they lower themselves and sit on the floor. Near the two-thirds point, the same person rises to stand again and remains standing through the end. The camera stays fixed without cuts or other people.''
\item \emph{State transition with a persistent entity or setting} (\texttt{pst-101\_person\_dog\_stand\_sit}).
$E=$ person; $A=$ standing dog; $B=$ sitting dog.

``A person and one standing dog share a clear full-body studio shot. The person stays in the same spot throughout the video. Around halfway through, the dog lowers into a seated pose and remains sitting beside the person until the end. The camera does not move and no additional subjects enter.''
\item \emph{Trigger-conditioned state transition} (\texttt{trg-041\_\allowbreak person\_\allowbreak triggers\_\allowbreak dog\_\allowbreak standing}).
$E=$ person; $A=$ sitting dog; $B=$ standing dog.

``A continuous full-body shot begins with one dog sitting alone in a quiet room. The dog remains seated for roughly the first third of the video. About one-third of the way through, a person walks into the frame and approaches while the dog remains seated. Around two-thirds of the way through, the dog responds to the person by standing up, and the person remains nearby through the end. The same dog and person stay in frame, with no cuts, duplicate subjects, or camera movement.''
\item \emph{Alternative final state} (\texttt{afs-161\_dog\_sit\_to\_stand\_or\_lie}).
$A=$ sitting dog; $B=$ standing dog; $C=$ lying dog.

``One dog sits upright in a clear full-body studio view for the opening third. During the middle third, it either rises onto all four paws or lowers into a lying pose, but it does not perform both actions. By the start of the final third, the dog is either standing or lying and remains in that selected pose through the end.''
\item \emph{Anchored entity choice} (\texttt{aec-141\_tree\_dog\_or\_cat}).
$E=$ tree; $A=$ dog; $B=$ cat.

``A single tree remains centered and alone throughout the opening third of a fixed garden view. During the middle third, either one dog may walk in or one cat may walk in, but not both. Whichever animal arrives stays near the tree throughout the final third without any duplicate animals.''
\end{enumerate}

\paragraph{Evaluation specifications.}
We assign each sampled frame its timestamp and use half-open
phase intervals, except that the final phase includes the video
endpoint. We evaluate temporal adherence using a phase-based
criterion. Let $t_i=i\Delta$ denote the sampled timestamps and
let $r=\lceil\delta/\Delta\rceil$, where $\delta=0.5$ seconds.
A phase predicate $p$ satisfies the persistence requirement
within interval $J$ if
\[
W_J(p)=1
\quad\Longleftrightarrow\quad
\exists i:\;
\{t_i,\ldots,t_{i+r}\}\subseteq J
\;\land\;
\bigwedge_{k=0}^{r} p(v_{i+k}),
\]
where the witness must use valid sampled-frame indices.
Thus, $p$ must hold at $r+1$ consecutive sampled frames whose
first-to-last timestamp span is at least $\delta$, with the
entire run contained in $J$.

The predicate $p$ jointly encodes the required presence and
absence of visual concepts. Explicit persistence requirements
are evaluated over their stated intervals rather than reduced
to a $0.5$-second witness. We use the same duration threshold
across the evaluated 5--6-second videos to require sustained
evidence beyond an isolated frame while allowing flexibility
in transition timing.

\textit{Interpretation of the criterion.} We evaluate whether each required phase occurs within its designated interval and persists for at least 0.5 seconds. We intentionally leave the intervening transitions unconstrained: intermediate configurations may not map unambiguously to the discrete propositions. For example, a partially open door may be labeled either open or closed while it moves between states. Requiring a sustained witness for each phase establishes that the requested states occur in the specified temporal order without prescribing the duration or frame-by-frame labeling of the transition between them. This criterion does not require each phase to occupy its entire interval or restrict behavior outside the witness runs, except where the specification imposes explicit persistence or absence constraints.

\section{Experimental setup and evaluation}
\label{appendix:experimental-setup}

\subsection{Models and sampling settings}
\label{appendix:models}
We evaluate TSG on the 220 prompt--specification pairs described in Appendix~\ref{appendix:benchmark}. Table~\ref{tab:experimental-models} summarizes the generation settings. Resolutions are given as height $\times$ width. We define duration as the time between the first and last decoded frames, $T=(F-1)/f$, where $F$ is the frame count and $f$ is the frame rate. Thus the Wan clips span five seconds and the CogVideoX clips span six seconds. Fractional specification intervals are compiled for each model's actual duration and temporal grid.

\begin{table}[t]
\centering
\small
\begin{tabular}{@{}lrrrrl@{}}
\hline
Model & Resolution & Frames/FPS & Latent frames & Steps & CFG \\
\hline
Wan 2.1 1.3B & $480\times832$ & $81/16$ & 21 & 50 & 6 \\
Wan 2.2 TI2V 5B & $704\times1280$ & $121/24$ & 31 & 50 & 5 \\
Wan 2.2 T2V A14B & $704\times1280$ & $81/16$ & 21 & 40 & 4 / 3 \\
CogVideoX 2B & $480\times720$ & $49/8$ & 13 & 50 & 6 \\
CogVideoX 5B & $480\times720$ & $49/8$ & 13 & 50 & dynamic (6) \\
\hline
\end{tabular}
\caption{Generation settings for the TSG benchmark. Wan 2.2 A14B uses CFG scales 4 and 3 for its high- and low-noise experts, respectively. CogVideoX 5B uses its dynamic CFG schedule with scale parameter 6. All methods compared within a model share these settings. The TempoControl comparison uses the Wan 2.1 row.}
\label{tab:experimental-models}
\end{table}

We use the diffusers library to obtain model implementations. The checkpoints are \texttt{Wan-AI/Wan2.1-T2V-1.3B-Diffusers}, \texttt{Wan-AI/Wan2.2-TI2V-5B-Diffusers}, \texttt{Wan-AI/Wan2.2-T2V-A14B-Diffusers}, \texttt{THUDM/CogVideoX-2b}, and \texttt{THUDM/CogVideoX-5b}. The TI2V checkpoint is used in text-to-video mode. Wan uses \texttt{UniPCMultistepScheduler}, with flow shift 3 for Wan 2.1 and 5 for both Wan 2.2 models. The fused prediction is passed to the scheduler once per denoising step, preserving its multistep state. CogVideoX 2B uses \texttt{CogVideoXDDIMScheduler}; CogVideoX 5B uses \texttt{CogVideoXDPMScheduler}. Both use trailing timestep spacing. For CogVideoX 5B, the implemented dynamic scale is
\[
 g(t)=1+6\,\frac{1-\cos\!\left(\pi\left((N-t)/N\right)^5\right)}{2},
\]
where $N=50$ and $t$ is the scheduler timestep value, not the zero-based sampling-step index. Other CFG scales remain constant, except for the expert-dependent A14B scales.

For each model and method, we generate one video per prompt. The TSG benchmark seed is $5000+j$, where $j$ is the numeric component of the example identifier. Each base run saves its initial latent tensor; the paired TSG run reloads that tensor, with shape, dtype, and SHA-256 checks. CogVideoX 5B additionally records and replays the sampling-generator state used by its stochastic DPM sampler. Prompts and negative prompts are identical within each paired comparison. The Wan negative prompt on this benchmark is: ``bright tones, overexposed, static, blurred details, subtitles, text, watermark, low quality, worst quality, jpeg artifacts, deformed''. CogVideoX uses no negative prompt. 

\subsection{TSG implementation and hyperparameters}
\label{appendix:tsg-settings}
The main TSG benchmark uses online attention steering with norm ratio $\rho=1$, applied during the first 15 denoising steps (indices 0--14). Ordinary CFG remains active throughout sampling. We set the additional weight to zero when the residual norm is at most $10^{-12}$. We use the z-score--sigmoid calibration with temperature
$\tau = 1.0$ and denominator stabilizer $\epsilon = 10^{-6}$, holding
the temporal mean and standard deviation fixed during differentiation.
After broadcasting and centering the attention-logit biases, we clip
them to $[b_{\min}, b_{\max}] = [-1.0, 1.0]$. We use bias step size \(\eta=1\).
We steer all attention heads in all transformer layers: 30 layers for Wan 2.1, Wan 2.2 TI2V, and CogVideoX 2B; 40 per active expert for Wan 2.2 A14B; and 42 for CogVideoX 5B.

\subsection{TempoControl comparison}
\label{appendix:tempocontrol}

We use all 80 one-object and 82 two-object prompts from the official TempoControl benchmark repository. One-object prompts request an initially absent object that appears during the second, third, fourth, or last second and remains visible, with 20 prompts for each onset. Two-object prompts require a static object throughout, with another object appearing in the second half. Binary masks are derived from the requested timing and the actual video frame rate and duration, rather than reusing the original 21-bin masks at a different duration. For our five-second Wan clips, the four one-object onsets occur at sampled indices 4, 8, 12, and 16, respectively; the object must be absent before its onset and present from that index through index 20. Two-object masks require absence of the entering object at indices 0--9 and presence at indices 10--20, together with the static object throughout.

To obtain TSG specifications, we convert each maximal constant segment of a binary mask into an always requirement for the proposition or its negation and conjoin the segment requirements. The static-object requirement is a global always formula. The runtime steering grid with Wan2.1 1.3B has 21 latent frames. All four comparison methods, i.e., CFG-only, TempoControl, one-round TempoControl, and TSG, use Wan 2.1 at $480\times832$, 81 frames, 16 FPS, 50 steps, CFG 6, and flow shift 3. We use seed 42 and replay the initial latent tensor saved by the full TempoControl run for each prompt.

Our TempoControl port preserves the released optimization order using the Diffusers backbone: latent optimization follows the scheduler update and queries attention at the preceding timestep. It optimizes the latent with AdamW (weight decay 0.01) at steps 0--10, with learning rate $5\times10^{-4}$ for one-object and $10^{-3}$ for two-object prompts. The objective uses energy weight 0.3 and entropy weight 10, with query chunks of 512 positions and activation checkpointing for the attention forward pass. The full variant allows up to ten optimization updates per guided step, following official recommendations in the repository. Early stopping checks whether the first controlled concept's Pearson correlation with its target mask exceeds 0.9, before applying the next update. The one-round variant changes only the maximum update count to one; it retains this early-stopping check and the same 11 guided steps.

\subsection{Ablation protocol and parameter selection}
\label{appendix:ablations}
The guidance-strength ablations in Section~\ref{sec:discussion} compare Wan 2.1 and Wan 2.2 TI2V on the same 25 samples. Every configuration replays its model's saved base latent tensor for each example. We retain the model-specific settings in Table~\ref{tab:experimental-models}, all attention layers and heads, the first 15 guided steps, step size 1, and the calibration and clipping settings above. The additional strengths are $\rho\in\{2,3,4,5\}$, compared with the existing base and $\rho=1$ runs. Only $\rho$ changes within this sweep.

The CFG interaction ablations also use the same base settings but modify the sampling as required. For the double strength CFG runs, we double the CFG strengths in Table~\ref{tab:experimental-models} for Wan2.1 and Wan2.2 5B. For TSG-only, we run TSG for all 50 steps and replace the conditional directions in CFG's update  with the TSG predicted specification-guided direction (CFG's guidance scale is used).

\subsection{Compute and measurement protocol}
\label{appendix:compute}
Wan 2.1, Wan 2.2 TI2V, both CogVideoX models, and the TempoControl comparison are run on a cluster with NVIDIA B200 GPUs. The Wan 2.2 A14B model is run on a cluster with NVIDIA H200s. The models use bfloat16 except CogVideoX 2B, which uses float16; the VAE uses float32. The recorded runtime is PyTorch 2.9.1 with CUDA 12.8, Diffusers 0.36.0, and Transformers 4.57.3. CogVideoX enables VAE tiling and slicing. The attention implementation uses PyTorch scaled-dot-product attention for ordinary computation and explicit attention-score computation where needed for capture and steering; the exact kernel selected by the runtime is not recorded in the per-video profiles.

Timing uses a monotonic clock with CUDA synchronization at measurement boundaries. Before each measured video, the profiler empties the CUDA allocator cache and resets peak-memory statistics after model loading. There is no separate discarded warm-up video in this measurement procedure. Generation latency includes initial-noise preparation or replay, prompt encoding, denoising and guidance, and VAE decoding. It excludes model loading, video export, evaluation, and separately timed diagnostic artifacts. Diagnostic memory allocations can still contribute to the recorded peak. When reported as mean $\pm$ standard deviation, the standard deviation is the sample standard deviation across videos, not a confidence interval or variation over repeated seeds. Peak memory is the maximum of the per-video PyTorch allocated-memory peaks, including resident model weights, converted to GiB; it is not device-wide memory usage or allocator-reserved memory. Hardware differs for A14B, so cross-model latency comparisons also reflect that difference.

\subsection{Evaluation metrics}
\label{appendix:metrics}

\paragraph{Frame sampling and specification accuracy.}
We evaluate 21 uniformly spaced decoded frames for every model, including the first and last frame. For a video with $F$ frames, sampled index $i$ is $\lfloor i(F-1)/20\rfloor$, for $i=0,\ldots,20$. 
We extract binary concept traces using \texttt{Qwen/Qwen2.5-VL-7B-Instruct}.  \emph{Specification satisfaction accuracy (Sat Acc.)} evaluates the prompt-derived specification under the Boolean semantics from Appendix~\ref{appendix:mtl_semantics}. 

\paragraph{Visual quality.}
We use the MUSIQ-SPAQ model underlying VBench's imaging-quality dimension~\citep{vbench}. It scores the same sampled frames used for specification evaluation, resizing the longer side to at most 512 pixels according to the VBench preprocessing. We average over frames within each video and then over videos. The table reports raw MUSIQ units, with larger values indicating better perceptual frame quality. Latency and peak GPU memory follow the measurement definitions above.

\paragraph{Evaluator configuration.}
The Qwen evaluator makes one request per sampled frame, asking for all relevant propositions together. The frame is converted to a uint8 PIL image and passed through the model's image processor and chat template. The application sends one user message containing the image and the verification prompt. Decoding is greedy, with at most 256 new tokens and one frame per request. The evaluator sees proposition definitions but not the requested temporal schedule or neighboring frames.

The fixed user-prompt preamble is reproduced below; the schema-specific field definitions are then appended from each example's proposition contexts.
\begin{quote}\small
Inspect only the supplied frame. Report what is visibly supported, not what ought to happen.
The scene may be stylized or synthetic. Do not infer missing objects or a desired temporal phase.
Return only one JSON object with exactly the keys described below, without explanations.

Object-presence fields: ``present'' means any recognizable visible part, even at the edge,
small, or partially occluded. ``absent'' means confidently not visible. ``unclear'' means
the evidence cannot establish presence or absence; do not equate uncertainty with absence.

State fields: return a JSON list of the exact state names visibly supported. Normally
one state applies. If multiple states visibly coexist, list all of them; do not force
a valid phase. Return [] only when you can confidently establish that none applies.
Return [``unclear''] if the state cannot be determined, the transition is ambiguous,
or the relevant entity is not recognizable. Do not combine ``unclear'' with other labels.
Use the same visible entity when distinguishing its states. A closed door is shut;
an open or ajar door is not closed. If the opening cannot be seen clearly, use unclear.
Standing and sitting refer to the actual pose; an ambiguous intermediate pose is unclear.
For traffic lights, judge the illuminated signal, not an unlit colored lens.
\end{quote}
For example, a pose-transition request appends a \texttt{state} field listing the allowed compound state labels and their definitions. Entity-presence fields instead allow \texttt{present}, \texttt{absent}, or \texttt{unclear}. City-lighting alternatives additionally receive explicit daytime, sunset, and nighttime definitions. 

Parsing accepts a JSON object, optionally within a JSON code fence, and normalizes case, whitespace, underscores, and Unicode dash separators in labels. Multiple supported states may be returned simultaneously; we do not force the answer to match a valid temporal phase. Explicit unclear labels and omitted fields map to false in the binary trace. Unknown fields are ignored; invalid known fields, unknown state labels, and inconsistent aliases raise protocol errors. The current evaluator allows two schema-correction retries on the same frame and preserves the invalid responses; exhausted retries leave an error record rather than a fabricated verdict. Earlier saved evaluations without retries remain identified by their raw-response records and source snapshots.

Across all five models, Qwen returned ``unclear” labels for at least one proposition in 433 of 46,200 individual frames (0.94\%). We assessed sensitivity to these labels using conservative and optimistic treatments across all five models. Conservatively, unclear labels support neither positive nor negative literals; \textit{no currently passing video fails under this treatment}, leaving reported accuracies unchanged. Optimistically, unclear labels may take Boolean values that satisfy the specification. With models weighted equally, this increases CFG accuracy by \(0.73\) percentage points and TSG accuracy by \(1.36\) percentage points.

\paragraph{TempoControl object and action metrics.}
Object evaluation uses the \texttt{YOLOv10x} detector with Ultralytics 8.4.152 on 21 sampled frames. The detector is invoked with its prediction defaults. Detected class names and requested labels are compared literally after lowercasing.

For one-object clips, a sampled frame is correct if detected presence matches the target mask. For two-object clips, correctness additionally requires the static object in both phases: before entry, the static object must be detected and the entering object absent; after entry, both must be detected. Temporal accuracy is the mean per-video fraction of correct frames. This frame-wise metric differs from the whole-video specification accuracy used on the TSG benchmark. 

\end{document}